\documentclass[10pt,journal,compsoc]{IEEEtran}

\ifCLASSOPTIONcompsoc
  \usepackage[nocompress]{cite}
\else
  \usepackage{cite}
\fi
\usepackage{xcolor}
\usepackage{amsmath}
\usepackage{amssymb}
\usepackage{booktabs}
\usepackage{algorithm}
\usepackage{multirow}
\usepackage{algorithmic}
\usepackage{graphicx}
\usepackage{bm}
\usepackage{color}
\usepackage{url}
\usepackage{float}
\usepackage{subcaption}

\usepackage[numbers,sort&compress]{natbib}

\usepackage[pagebackref,breaklinks,colorlinks]{hyperref}
\usepackage{makecell}
\usepackage{xspace}

\usepackage{stfloats}

\newcommand{\wjj}[1]{{\color{blue}#1}}
\newcommand{\wj}[1]{{#1}}

\makeatletter
\DeclareRobustCommand\onedot{\futurelet\@let@token\@onedot}
\def\@onedot{\ifx\@let@token.\else.\null\fi\xspace}

\def\eg{\emph{e.g}\onedot} 
\def\ie{\emph{i.e}\onedot}

\makeatother

\ifCLASSINFOpdf
\else
\fi

\begin{document}

\title{CPR++: Object Localization via Single Coarse Point Supervision}

\author{Xuehui Yu, Pengfei Chen, Kuiran Wang, Xumeng Han, Guorong Li,~\IEEEmembership{Senior Member,~IEEE,}  Zhenjun Han,~\IEEEmembership{Member,~IEEE,}  Qixiang Ye,~\IEEEmembership{Senior Member,~IEEE,}  Jianbin Jiao,~\IEEEmembership{Member,~IEEE} 


\IEEEcompsocitemizethanks{\IEEEcompsocthanksitem Corresponding author: Zhenjun Han.}
\IEEEcompsocitemizethanks{\IEEEcompsocthanksitem X. Yu, P. Chen, K. Wang, X. Han, Z. Han, Q. Ye, J. jiao  are with the School of Electronic, Electrical and Communication Engineering, University of Chinese Academy of Science (UCAS), Beijing, 100049, China. E-mail: \{yuxuehui17, chenpengfei20,  wangkuiran19, hanxumeng19\}@mails.ucas.ac.cn, \{hanzhj, qxye, jiaojb\}@ucas.ac.cn.}
\IEEEcompsocitemizethanks{\IEEEcompsocthanksitem G. Li is with the School of Computer Science and Technology, University of Chinese Academy of Science (UCAS), Beijing, 100049, China. E-mail: liguorong@ucas.ac.cn.}}


\markboth{Journal of \LaTeX\ Class Files,~Vol.~14, No.~8, August~2015}%
{Shell \MakeLowercase{\textit{et al.}}: }

\IEEEtitleabstractindextext{%
\begin{abstract}
Point-based object localization (POL), which pursues high-performance object sensing under low-cost data annotation, has attracted increased attention.
However, the point annotation mode inevitably introduces semantic variance due to the inconsistency of annotated points.
Existing POL heavily rely on strict annotation rules, which are difficult to define and apply, to handle the problem.
%
%
In this study, we propose coarse point refinement (CPR), which to our best knowledge is the first attempt to alleviate semantic variance from an algorithmic perspective.
CPR reduces the semantic variance by selecting a semantic centre point in a neighbourhood region to replace the initial annotated point.
Furthermore, We design a sampling region estimation module to dynamically compute a sampling region for each object and use a cascaded structure to achieve end-to-end optimization. 
We further integrate a variance regularization into the structure to concentrate the predicted scores, yielding CPR++.
We observe that CPR++ can obtain scale information and further reduce the semantic variance in a global region, thus guaranteeing high-performance object localization.
Extensive experiments on four challenging datasets validate the effectiveness of both CPR and CPR++. We hope our work can inspire more research on designing algorithms rather than annotation rules to address the semantic variance problem in POL. The dataset and code will be public at~\url{github.com/ucas-vg/PointTinyBenchmark}.

\end{abstract}

}

\maketitle

\IEEEdisplaynontitleabstractindextext

%
\IEEEpeerreviewmaketitle

\IEEEraisesectionheading{\section{Introduction}\label{sec:introduction}}

\IEEEPARstart{T}he core of the human visual system is the perception and recognition of objects in the line of sight. In computer vision, this is often formulated as drawing a bounding box around the object ~\cite{DBLP:journals/pami/LinGGHD20, DBLP:conf/nips/RenHGS15, DBLP:conf/eccv/LiuAESRFB16, DBLP:conf/cvpr/SunZJKXZTLYW021, DBLP:journals/pami/LiLWHXL22,DBLP:journals/pami/ChenLLSWZ21} or distinguishing the dense semantic content of the entire scene~\cite{he2017mask, huang2019ccnet, DBLP:journals/pami/BolyaZXL22, DBLP:journals/pami/HuSLLZZ22, DBLP:journals/pami/XieWDZL22, DBLP:journals/pami/WangZSKL22, DBLP:journals/pami/QiWCCZSJ22, DBLP:journals/pami/HuangWWLHS22, DBLP:journals/pami/OhLXK22, DBLP:journals/pami/FengLLWTYL22}.
However, such formulations require well-annotated images to learn perceptual models, which are often expensive or difficult to obtain.
Furthermore, in some real-world applications, we are only concerned with the position of an object instead of delineating it with a bounding box or segmentation mask.
For example, a robotic arm is aimed at a single location point to pick up an object~\cite{runow2020deep}.
This motivates research into point-based object localization (POL)~\cite{DBLP:conf/cvpr/RiberaGCD19, Song_2021_ICCV}. POL requires only simple and time-saving annotations, and specifically, it only needs the point-level object annotations to train a detector to locate objects with 2D coordinates in a given image.

\begin{figure*}[tb!]
\begin{center}
    \begin{tabular}{ccc}
    \includegraphics[width=0.95\linewidth]{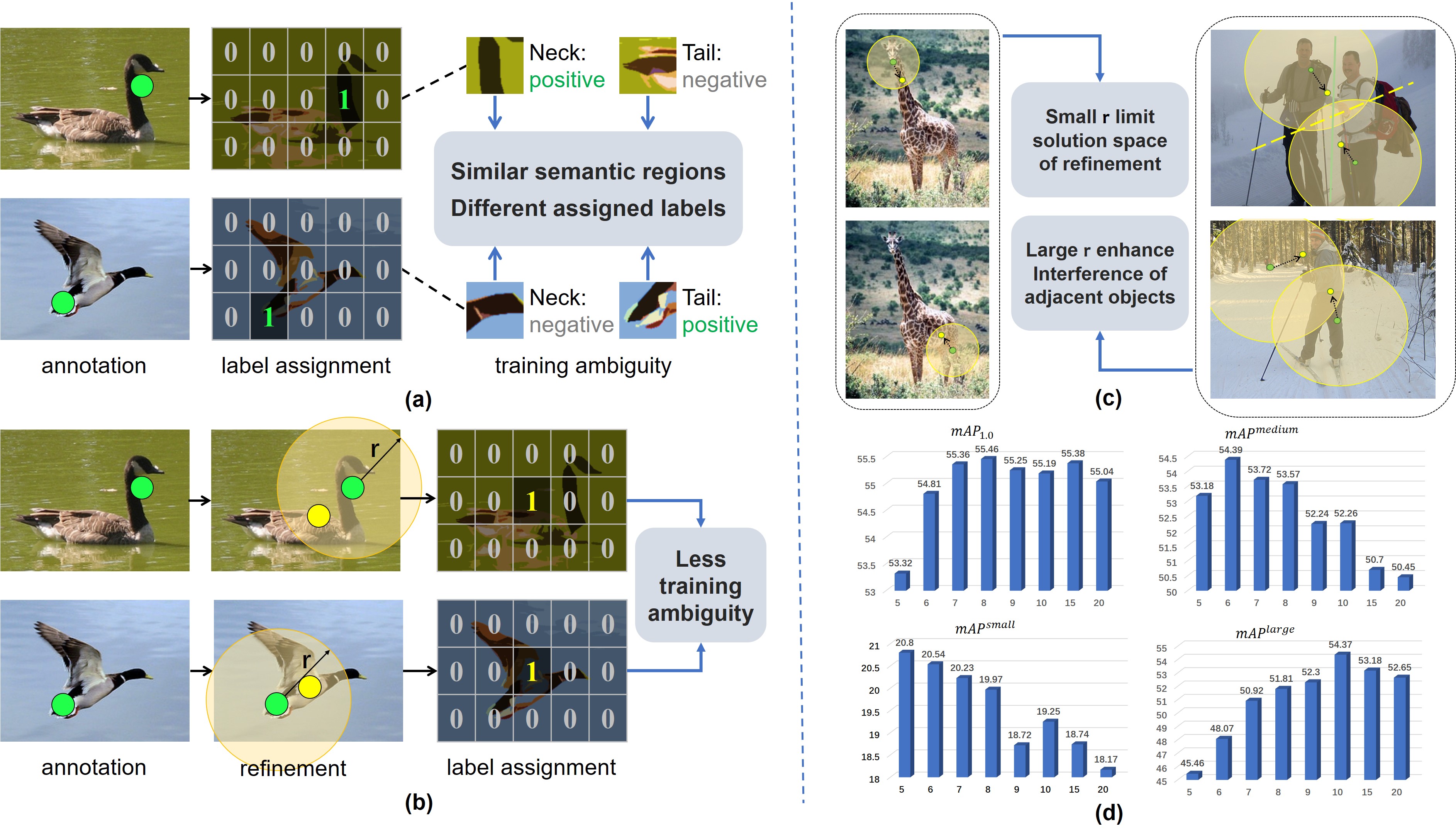}
    \end{tabular}
    \caption{The motivation of CPR/CPR++. (a) Examples of coarse point annotation and the problem of semantic variance. (b) CPR aims to find a semantic center point of the objects belonging to the same category to reduce training ambiguity. (c) Limitation of CPR. Due to the leak of scale information, either small or large $r$ has its problem. A small radius leads to the local semantic point instead of the global solution. Large radius results in merging with other object regions. (d) The performance of different sampling radii in CPR. When the radius gets larger, the ${\rm mAP}$ of large objects increases while that of small objects decreases. The performance drops when the radius is too large due to the interference of adjacent objects.}
    \vspace{-20pt}
\label{fig:motivation of ccpr}
\end{center}
\end{figure*}

There are multiple candidate points when annotating an object as a point. This leads to the problem of \textit{semantic variance}: 1) regions with different semantic information are labelled as positive samples of the class; 2) regions with similar semantic information are labelled differently.
Taking the \textit{bird} category as an example, we label the bird's different body parts (\eg, neck and tail) as positive based on the visible regions in the image. For different images in a dataset, the same body part of the bird (\eg, neck) may be labelled as positive and negative, respectively, as shown in Fig.~\ref{fig:motivation of ccpr}(a). As a result, during training, model treats the neck region as positive in one image and negative in another (the one where the tail is labelled). The semantic variance causes ambiguity and thereby confuses the model, leading to performance degradation.

\begin{figure}[tb!]
\centering
\subcaptionbox{Chairs with diff. shapes\label{fig:chair}}{
    \centering
    \includegraphics[height=0.30\linewidth]{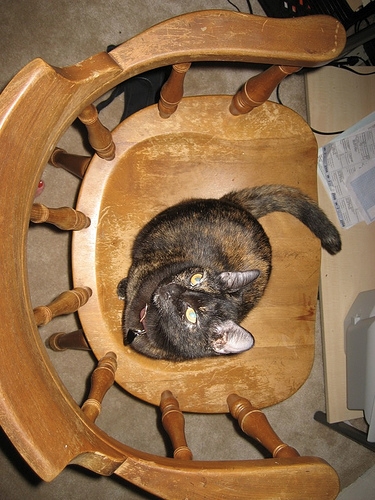}
    \includegraphics[height=0.30\linewidth]{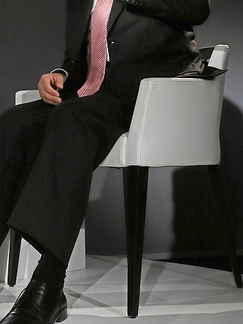}
}
\hfill
\subcaptionbox{Persons with diff. poses\label{fig:pose}}{
    \centering
    \includegraphics[height=0.30\linewidth]{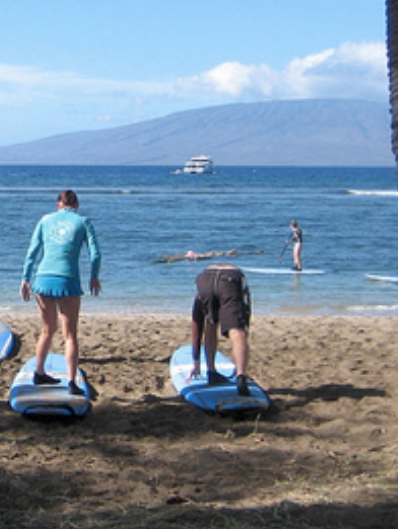}
    \includegraphics[height=0.30\linewidth]{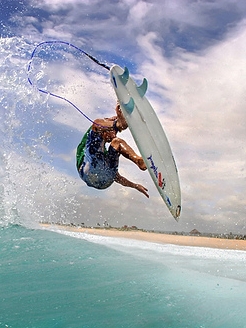}
}
\hfill
\subcaptionbox{Persons of diff. sizes\label{fig:scale}}{
    \centering
    \includegraphics[height=0.31\linewidth]{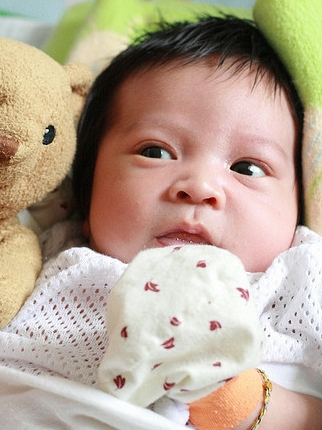}
    \includegraphics[height=0.31\linewidth]{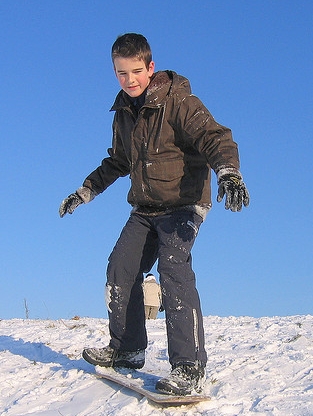}
    \includegraphics[height=0.31\linewidth]{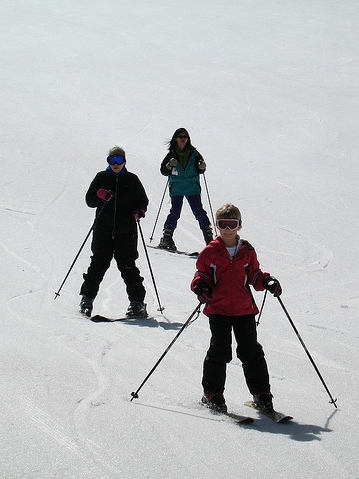}
    \includegraphics[height=0.31\linewidth]{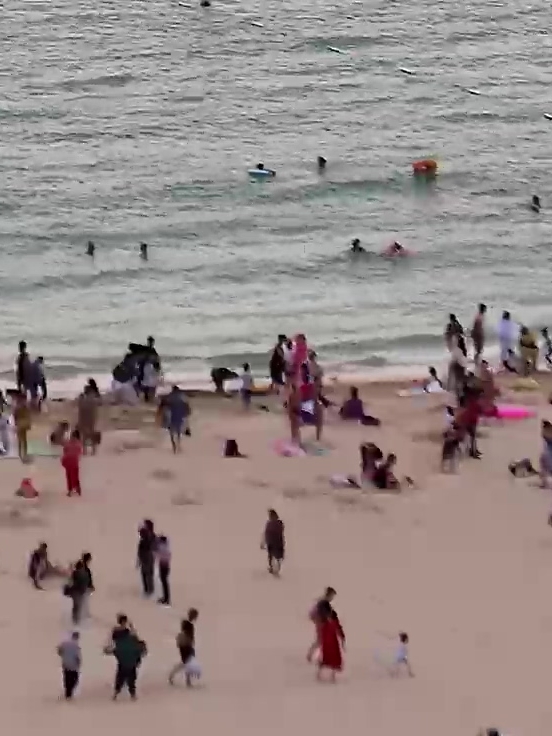}
}
\setlength{\belowcaptionskip}{-0.5cm}
  \caption{Difficulty of key-point-based annotation. (a) Key points are hard to define due to the large in-class variance of shape. (b) Key point (\eg, head) does not exist due to multiple poses and views. (c) The key point's granularity (\eg, eye, forehead, head, and body) is hard to determine due to multiple scales.}
\label{fig:key point problem}
\end{figure}

Existing works~\cite{Song_2021_ICCV,DBLP:journals/pami/WangGLL21} have attempted to mitigate this issue by setting strict annotation rules, such as annotating the pre-defined key-point areas of the object. 
However, this practice suffers from the following challenges.
1) Key points are not easy to define, especially for some broadly defined categories in which they do not have a specific shape (Fig.~\ref{fig:key point problem} (a)).
2) Key points may not exist in the image due to the different poses of the objects and different camera views (Fig.~\ref{fig:key point problem} (b)).
3)~When objects have large-scale variance, it is challenging to decide the appropriate granularity of the key points (Fig.~\ref{fig:key point problem} (c)). For example, if the head of a person is defined as a key point (coarse-grained)~\cite{Song_2021_ICCV}, there remains a large semantic variance for the large-scale person instances (\eg, whether to annotate the eye or the nose). If the eye is labelled as a key point (fine-grained)~\cite{DBLP:journals/pami/WangGLL21}, the position of the eyes for a small-scale person instance cannot be identified. 
The above challenges often exist in real-world object datasets (\eg, COCO and DOTA).
As such, the complicated annotation rules cannot satisfactorily solve the semantic variance problem. Instead they increase the annotation difficulty, which might limit the application scope of previous POL methods.

In this work, we propose a coarse point-based localization (CPL) paradigm, that enables us to train a general and accurate object localizer. 
As shown in Fig.~\ref{fig:pipeline}, we first adopt a coarse point annotation strategy to label \textit{any} point on an object. Then a simple but effective algorithm named coarse point refinement (\textbf{CPR}) is proposed to refine the coarse points during training. Finally, the refined points are used as supervision to train a localizer. 
Our CPR is the first attempt to alleviate semantic variance from an algorithmic rather than  an annotation perspective.
CPR finds the semantic points around the annotated point through multiple instance learning (MIL) ~\cite{DBLP:journals/ai/DietterichLL97} and then weights the semantic points on average to obtain the semantic centre, which has a smaller semantic variance and a higher tolerance for localization errors.

To further reduce the annotation semantic variance, we also consider the \textit{sampling range} of semantic points.
Due to the lack of object scale information in POL annotations, CPR sets a fixed sampling range for all objects. That is, the semantic points are searched from the $r$-circle neighbour region of the initial annotation points.
The fixed sampling range of $r$-circle may not be suitable in some cases.
For example, for adjacent objects, setting a large $r$ will make the sampling area of different objects overlap, which might confuse the training and lead to localization mismatch as shown in Fig.~\ref{fig:motivation of ccpr}(c). 
This motivates us to explore a dynamic sampling range for each object. To this end, we further design \textbf{CPR++} that progressively generates a dynamic sampling region for each object. 
%
Based on semantic points produced by CPR,
a sampling region estimation module in CPR++ is designed to measure these semantic points' spatial uncertainty. The spatial uncertainty is related to the scale of the object and thus can be used to decide the adaptive sampling region. Then the adaptive sampling region is used to perform a new CPR.
To achieve end-to-end optimization, we train different CPR as different stages in a cascade fashion. 
In addition, to make loose predictzongshued scores more concentrated to a single point of each object, variance regularization is included into CPR++ to constrain the training procedure, achieving a better optimization.
The newly introduced strategies allow CPR++ to further reduce the semantic variance across images, resulting in a high-performance object localizer.

The contributions of this paper are as follows:

1) We provide an in-depth investigation of the POL and propose a coarse point-based localization (CPL) paradigm for generic object localization, extending the previous works to a multi-class and multi-scale POL task.

2) The proposed CPR approach utilizes multiple instance learning (MIL) to search the semantic center point from the $r$-circle neighbour region of the initial annotation points, mitigating the semantic variance from an algorithmic perspective rather than rigid annotation rules.

3) We propose sampling region estimation module and a cascade structure to alleviate the dilemma of setting sampling ranges, resulting in CPR++ that adaptively acquires dynamic sampling regions from coarse point annotations. 
Meanwhile, CPR++ introduces variance regularization to reduce the semantic variance further.

4) The experimental results show our CPR++ is effective for CPL, obtains a comparable performance with the center point (approximate key point) based object localization on COCO dataset, and improves the performance over 10 points compared with the baseline. 

This article extends from our conference paper~\cite{DBLP:CPR}. The major extensions include the following: 1) We provide more ablation study of CPR in Sec.~\ref{Sec: Ablation Studies on CPR} and a detailed analysis of the limitations and dilemmas caused by the fixed sampling radius setting for CPR in Sec.~\ref{sec: cascade designing}. 2) We give the description of CPR++ in Sec.~\ref{sec:cpr++} and demonstrate the structure of CPR++, which adaptively generates dynamic sampling regions for each object by sampling region estimation module and continuously optimizes the semantic center points in a cascade fashion in Sec.~\ref{sec: cascade designing}. 3) The newly introduced variance regularization, as described in Sec.~\ref{sec:variance-loss}, further reduces the semantic variance of the coarse point annotation. 4) The detail ablation studies on CPR++ are given in Sec.~\ref{sec: Experimental analysis on CPR++}. 5) A visualization analysis of CPR and CPR++ are given in Sec.~\ref{sec: visualization analysis}. 6) The more detail performances of different metric are given in Sec.~\ref{sec: Comparison with the State-of-the-Art}. Further, more dataset is used to validate the efficiency of our proposed CPR and CPR++.


\section{Related Work}

In this section, we review relevant point-based vision tasks and vision tasks with multiple instance learning, cascade structure and uncertainty estimation. 

\subsection{Vision Tasks under Point Supervision} 
\textbf{Pose Estimation.} 
Human pose estimation~\cite{DBLP:journals/pami/CaoHSWS21} aims to locate the position of joint points accurately. 
There are several benchmarks built for the task, \eg, COCO~\cite{lin2014microsoft} and the Human3.6M ~\cite{DBLP:journals/pami/IonescuPOS14} datasets are the most well-known ones for 2D and 3D pose estimation.
In these datasets, annotations are a set of accurate key points, and the predicted results are human poses rather than the location of persons.


\textbf{Crowd Counting.} In this task, accurate head annotation is utilized as point supervision~\cite{DBLP:conf/cvpr/ZhangZCGM16, DBLP:journals/pami/WangGLL21, Song_2021_ICCV}. The crowd density map~\cite{DBLP:conf/nips/LempitskyZ10, DBLP:conf/cvpr/JiangZXZLZYP20,DBLP:conf/eccv/HuJLZHCD20}, generated by head annotation, is chosen as the optimization objective of the network. Furthermore, crowd counting focuses on the number of people rather than each person's position. It depends on precise key points such as the human head. Nevertheless, the coarse point object localization task only requires coarse position annotation on the human body.

\textbf{Object Localization.} 
\cite{DBLP:p2bnet} conducts object detection via single quasi-center point annotation. \cite{BSETIE, WISE-Net} use point annotation to supervise instance segmentation. 
Different from object detection~\cite{DBLP:conf/iccv/LinGGHD17, DBLP:conf/nips/RenHGS15, DBLP:conf/eccv/LiuAESRFB16, DBLP:conf/cvpr/SunZJKXZTLYW021, DBLP:fusionfactor, DBLP:journals/pami/ZhangWLJY22}, the bounding box is over-modeling for object localization applications~\cite{DBLP:conf/cvpr/RiberaGCD19} that are not interested in the object's scale.  \cite{DBLP:conf/cvpr/RiberaGCD19, Song_2021_ICCV} train a localizer and predict object locations with points instead of bounding boxes. These tasks are summarized as POL in our paper. However, they heavily rely on key-point annotations to reduce semantic variance.

Different from the above-mentioned tasks, our CPL relies upon a coarse point instead of key points and deals the semantic variance problem with a novel approach.
\subsection{Vision Tasks with Multiple Instance Learning}
The paradigm of MIL~\cite{DBLP:journals/ai/DietterichLL97} is that a bag is positively labeled if it contains at least one positive instance; otherwise, it is labeled as a negative bag. 
Inspired by weakly supervised object detection tasks, the proposed CPR method follows the MIL paradigm. With the object category and the coarse point annotation, we consider sampled points around each annotated point as a bag and utilize MIL for training. 

\textbf{Image-level Tasks.} An image is divided into patches, where patches are seemed as instances and the entire image as a bag. Content-based image retrieval~\cite{DBLP:journals/pr/ZhangWSZ10,DBLP:conf/icml/ZhangGYF02} is a conventional MIL task, which just classifies images by their content. 
If the image contains at least one object of a class, the whole bag can be seen as a positive sample for that class.
Otherwise, the bag will be regarded as a negative sample.

\textbf{Video-level Tasks.} Firstly, the video is divided into segments, which will be classified separately and then the whole video is seemed as a bag. Following the above pre-processing, MIL is used to identify specific events in videos~\cite{DBLP:conf/cvpr/FengHZ21,DBLP:conf/cvpr/NguyenLPH18,DBLP:conf/cvpr/SultaniCS18}. Additionally, some researchers have applied MIL to video target tracking~\cite{DBLP:conf/cvpr/BabenkoYB09,DBLP:journals/pami/BabenkoYB11}. A pre-classifier is trained to identify and track an object. Then, this classifier generates candidate boxes and views them as a bag to train a MIL classifier.

\textbf{Object-level Tasks.}
MIL is widely used in weakly supervised object localization and detection (WSOL~\cite{DBLP:journals/pami/CinbisVS17} and WSOD~\cite{DBLP:conf/cvpr/BilenV16, DBLP:conf/ijcai/WangYZZ18,DBLP:conf/cvpr/WanLKJJY19, DBLP:journals/pami/TangWBSBLY20, DBLP:conf/cvpr/ChenFJC020, DBLP:journals/pami/WanWHJY19}), where only the image-level annotation is utilized. Firstly, Select Search~\cite{DBLP:conf/iccv/SandeUGS11}, Edge Box~\cite{DBLP:conf/eccv/ZitnickD14} or MCG~\cite{DBLP:MCG} methods are used to produce proposal boxes, which are then used as a bag and each of them as an instance. 
Finally, they classified positive and negative samples by judging whether the image contained at least one object of a specific class. WSOL/WSOD, only with image-level annotation, focus on local regions and can not distinguish instances due to the lack of object-level annotation. 


Annotation of CPL is a coarse point position and the category of each object. CPR views sampled points around the annotated point as a bag and trains object-level MIL to find a better and more stable semantic center.

\subsection{Vision Tasks with Cascade Structure}

In~\cite{DBLP:conf/cvpr/ViolaJ01}, the authors introduced Haar-like~\cite{DBLP:conf/iccv/PapageorgiouOP98} feature and utilized a boosted cascade strategy to realize fast face detection. Cascade CNN~\cite{li2015convolutional} uses three weak cascade classifiers to realize human face detection. Cascade R-CNN~\cite{DBLP:journals/pami/CaiV21} introduces cascade structure into Faster R-CNN~\cite{DBLP:conf/nips/RenHGS15} and becomes a classic object detector. Sever same detection heads are cascaded with different hyper-parameter settings like IoU threshold when positive and negative matching to achieve a coarse-to-fine effect. SparseRCNN~\cite{DBLP:conf/cvpr/SunZJKXZTLYW021} utilizes a cascade structure to replace RPN~\cite{DBLP:conf/nips/RenHGS15}. 
OICR~\cite{DBLP:oicr} introduces an iterative fashion into weakly supervised object detection and attempts to find the whole part instead of a discriminative part. WCCN~\cite{DBLP:wccn} introduces a class activation map to extract the best candidates for better MIL training in a cascade fashion. 
P2BNet~\cite{DBLP:p2bnet} introduces a cascade structure to iteratively refine predicted pseudo box from point annotation in point-supervised object detection task.

The cascade structure is used in this paper to solve the problem of imprecise sampling area under the condition of missing scale information.

\vspace{-8pt}
\section{Methodology}

\begin{figure}[tb!]
\begin{center}
    \begin{tabular}{ccc}
    \includegraphics[width=0.99\linewidth]{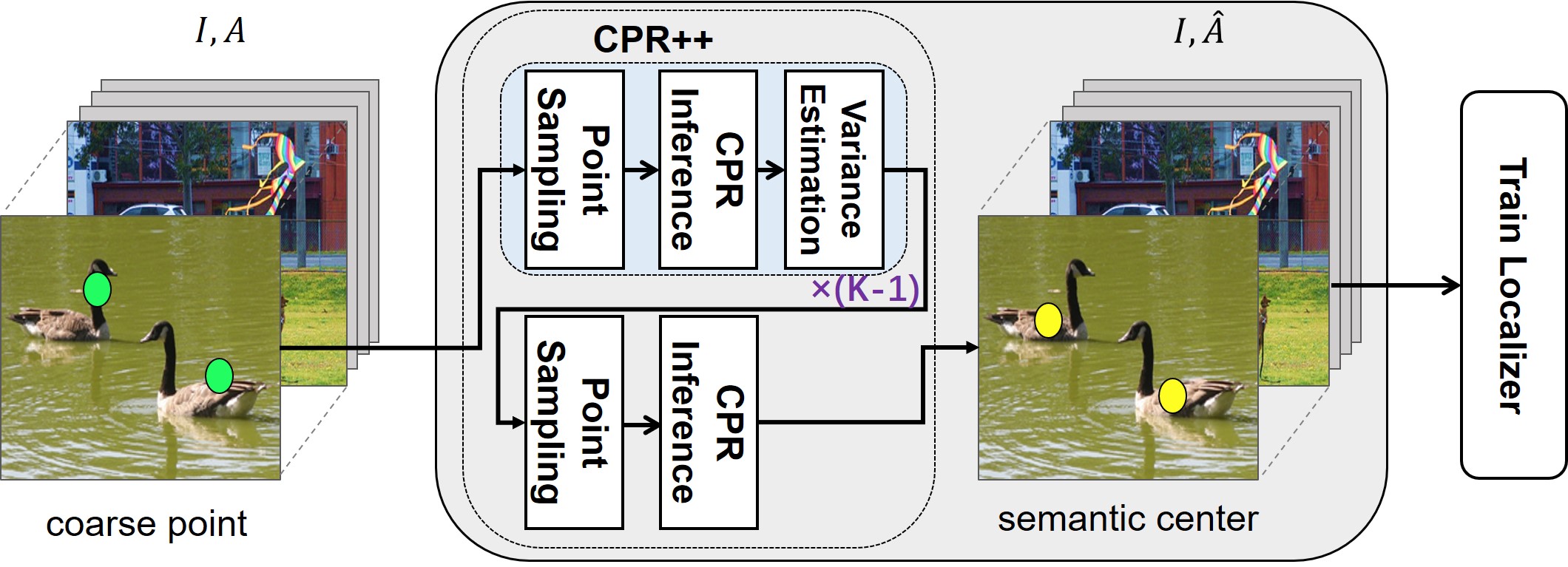}
    \end{tabular}
   \caption{Pipeline of coarse point-based localization (CPL). There are three steps: 1) Annotating objects as coarse points $A$; 2) Refining annotated points $A$ to semantic centers $\hat{A}$; 3) Training a localizer (\emph{e.g.}, P2PNet) with $\hat{A}$ as supervision. "×(K-1)" means the blue block repeats K-1 times and CPR++ becomes to CPR while K=1.
   }
\label{fig:pipeline}
\end{center}
\vspace{-15pt}
\end{figure}

\subsection{Overview}
Given a training dataset $D^{train}$ consisting of images and coarse point annotations of objects, the goal of this work is to learn a general and high-performance object localizer. 
To this end, we study a coarse point-based localization (CPL) paradigm, which consists of three steps (shown in Fig. \ref{fig:pipeline}). 1) Labeling objects with coarse points $A$; 2) Refining annotated coarse points $A$ to semantic centers $\hat{A}$; 3) Training a localizer (\emph{e.g.}, P2PNet) with $\hat{A}$ as supervision.

Our key focus is the second step of CPL. The semantic variance of annotated coarse points leads to training ambiguity and thus degrades localization performance. 
Therefore, instead of training a localizer directly using the initial point annotations, in Section \ref{sec:cpr}, we propose an effective method named coarse point refinement (\textbf{CPR}) to refine them into more accurate points with smaller semantic variance.
CPR can be reviewed as a \textit{pre-processing} that refines the initial coarse point annotation into a more conducive form for subsequent learning procedure of localizer.
To boost the point refinement of CPR, in Section \ref{sec:cpr++}, we further propose to progressively estimate the radius of each individual instance during the coarse point refinement, yielding CPR++.
In the following sections, we will introduce CPR first and then CPR++.

\begin{figure*}[tb!]
\begin{center}
    \begin{tabular}{ccc}
    \includegraphics[width=0.98\linewidth]{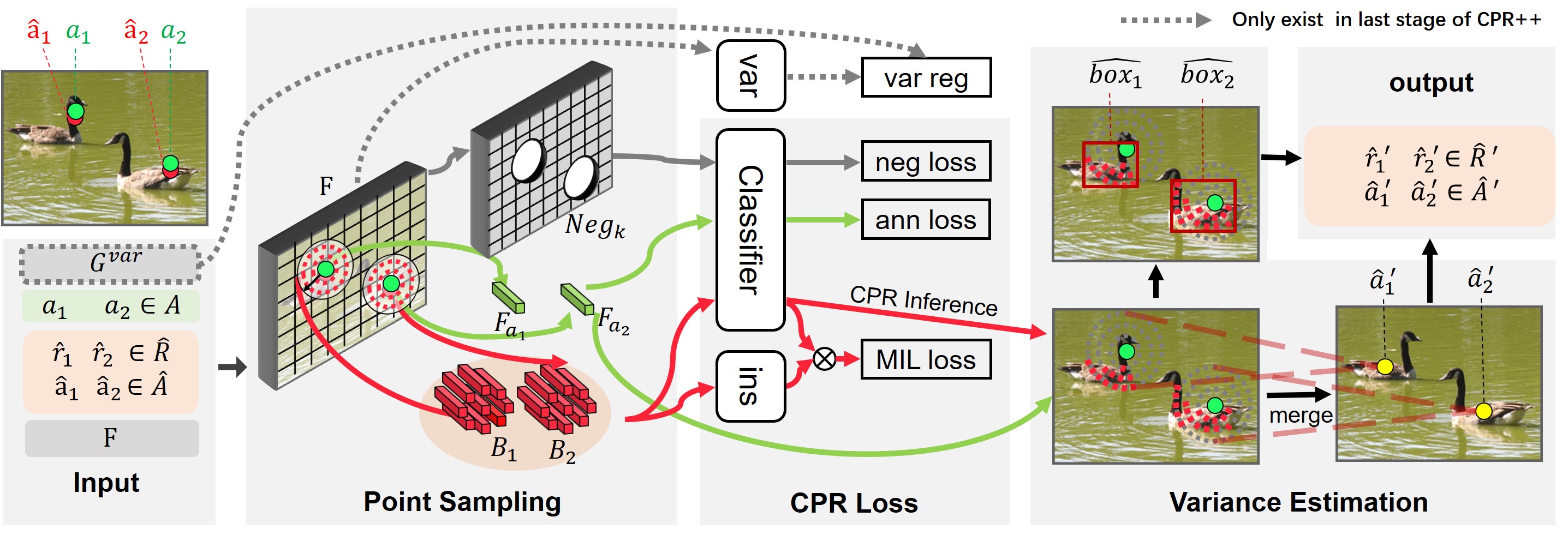}
    \end{tabular}
    \caption{
    The framework of CPR and sampling region estimation module of CPR++. Given feature map $F$, annotated point $A$ (green), center $\hat{A}$ (red) and radius $\hat{R}$ of point sampling, there are three steps in the CPR stage:
   1) Positive bags (\ie, $B_1, B_2$) and negative samples (\ie, $Neg_{k_c}$) are obtained by point sampling, and then feature vectors of these points are extracted on $F$. 2) Network is trained with the feature vectors based on CPR loss(MIL loss, annotation loss and negative loss) and variance loss. 3)
  Semantic points (red points on the birds) $B^+_1, B^+_2$ are selected by classification scores of points in the bag (\ie, $B_1, B_2$) predicted by the trained network (CPR Inference). 4) Finally, the refined points (yellow) $\hat{A}'$ are obtained by weighted averaging of the semantic points. For CPR++, the circumscribed rectangle of the semantic points is used to estimate the radius of the next stage (Sec.~\ref{sec: variance estimation}) and the variance map $G^{var}$ are fed into the network to conduct variance regularization (Sec.~\ref{sec:variance-loss}) for last CPR++ stage. 
  (Best viewed in color).
   }
\label{fig:method CPR}
\end{center} 
\vspace{-15pt}
\end{figure*}

\vspace{-3pt}
\subsection{CPR}\label{sec:cpr}

There are three key parts in CPR: 
1) Point Sampling: points in the neighborhood of the annotated point are sampled; 2) CPR Training: based on the sampled points, a network is trained to classify whether the points are in the same category as the annotated point; 3) CPR Inference:
based on the scores obtained by CPRNet and the constraints (details in Sec.~\ref{sec: CPRInfer}), the points, having similar semantic information with the annotated point, are chosen as the semantic points. For the origin CPR, these semantic points are weighted with their scores to obtain the semantic center as the refined point. 



\subsubsection{Point Sampling}\label{sec:Point sampler}
This section gives the formulation of point sampling, which need to do firstly in both $CPR\ training$ and $CPR\ inference$.
$K_c$ denotes the number of categories, $\hat{a}_j\in R^2$, $\hat{r}_j\in N^+$, and $C_j \in\{0, 1\}^{K_c}$ denote the sample center point's 2D coordinate, sample radius and the one-hot category label of $j$-th instance. $p=(p_x, p_y)$ denotes a point on a feature map.

\textbf{Point Bag Construction.}\label{sec: bag sampling} In Fig.~\ref{fig:method CPR}, to sample points uniformly in the neighborhood of $\hat{a}_j$, we define $\hat{r}_j$ circles with $\hat{a}_j$ as the center, where the radius of the $r$-th ($1 \leq r \leq \hat{r}_j$, $r\in N^+$) circle is set as $r$. Then we sample $r*u_0$ ($u_o$=8 by default) points with equal intervals around the circumference of the $r$-th circle, and obtain $Circle(\hat{a}_j, r)$. All sampled points of the $\hat{r}$ circles are defined as points' bag of $\hat{a}_j$, denoted as $B_j$ in Eq.~\ref{Eq:Bag_j}. The points outside the feature map are excluded.
\begin{equation}\small
\begin{aligned}
\textit{Circle}&(p, r)  = \bigg\{\bigg(p_x+r\cdot cos\left(2\pi \cdot \frac{i}{u_0 \cdot r}\right), p_y + \\
& r \cdot sin\left(2\pi\cdot\frac{i}{u_0\cdot r}\right)\bigg)\ |\  0 \leq i< r\cdot u_0, i\in N^+\bigg\}; \\
& B_j = \textit{bag\_sampling}(\hat{a}_j, \hat{r}_j) = \mathop{\cup}\limits_{1\leq r\leq \hat{r}_j}\ \textit{Circle}(\hat{a}_j, r).
\label{Eq:Bag_j}
\end{aligned}
\end{equation}

The sampling points $B_j$ are then used for calculating the MIL loss for CPRNet training and obtaining the semantic points for point refinement.

\textbf{Negative Point Sampling.}\label{sec: negative sampling} 
All integer points on the feature map that are not inside the sampling circles of all instances of a given category will be selected as negative samples. Specifically, the negative samples of category $k_c$ can be defined as:
\vspace{-5pt}
\begin{equation}\small
\begin{aligned}
Neg_{k_c} & = \{(p_x, p_y)| p_x\leq w, p_y \leq h, p_x \in N^+, p_y \in N^+\\
& \forall (\hat{a}_j, c_j)\ s.t. \ c_{jk_c}=1, ||p-\hat{a}_j|| > \hat{r}_j\},
\label{Eq:Neg_k}
\end{aligned}
\end{equation}where $||p-\hat{a}_j||$ is the Euclidean distance between $p$ and $\hat{a}_j$. $w$ and $h$ are the width and height of a given feature map.

\subsubsection{CPR Training}
This section gives the details of the objective function of training CPRNet based on the sampled points bag $B_j$ ($j \in \{1, 2, ..M\}$) and the negative points $Neg_{k_c}$ ($k_c \in \{1, 2, ..K_c\}$), where $M$ and $K_c$ are the amount of instances and categories. $U$ is defined as the number of points in $B_j$.

%
\textbf{CPRNet.}\label{sec:CPRNet architecture} CPRNet adopts FPN~\cite{DBLP:conf/cvpr/LinDGHHB17} with ResNet~\cite{DBLP:conf/cvpr/HeZRS16} as the backbone. Only P2 or P3 is used due to the lack of scale information in point annotation. After four 3$\times$3 conv layers followed by the ReLU ~\cite{DBLP:journals/jmlr/GlorotBB11} activation, the final feature map $\mathbf{F}\in \mathbb{R}^{h\times w \times d}$ is obtained, where $h\times w$ is 
the corresponding spatial size and $d$ is the dimension of channel. For a given point $p=(p_x, p_y)$, $\mathbf{F}_{p}\in \mathbb{R}^{d}$ denotes the feature vector of $p$ on $\mathbf{F}$. If $p$ is not an integer point, the bilinear interpolation is used to obtain $\mathbf{F}_p$.

\begin{algorithm}[tb!]
{
\small
\caption{CPR Training}
\label{Alg: CPRLoss}
\textbf{Input:} Annotated point $A$, Sampling center $\hat{A}$ and radius $\hat{R}$ of image $I$. \\
\textbf{Output:} training loss $L_{CPR}$.\\
\textbf{Note:} CPRNet $E$, bag sampling $BS$ and negative sampling $NS$ are described in Sec ~\ref{sec:Point sampler}. $\mathbf{F}$ is extracted feature map of image $I$  according to Sec. \ref{sec:CPRNet architecture}. \\
\vspace{-10pt}
\begin{algorithmic}[1]
\STATE \textit{// Step1: point sampling}
\STATE $B_j \leftarrow BS(a_j, \hat{a}_j, \hat{r}_j)$ for each $a_j \in A$, $\hat{a}_j \in \hat{A}$, $\hat{r}_j \in \hat{R}$, Eq.~\ref{Eq:Bag_j};
\STATE $Neg_{k_c} \leftarrow NS(k_c)$ for each category $k_c \in \{1,$ $2, ...K_c\}$, Eq.~\ref{Eq:Neg_k};
\STATE \textit{// Step2: loss calculating}
\STATE Calculate $L_{MIL}$ with $B_j$ and $\mathbf{F}$, Eq.~\ref{Eq:L_{bag}};
\STATE Calculate $L_{ann}$ with $A$ and $\mathbf{F}$, Eq.~\ref{Eq:S_{a_j}};
\STATE Calculate $L_{neg}$ with $Neg_{k_c}$ and $\mathbf{F}$, Eq.~\ref{Eq:S_{Neg}};
\STATE Sum $L_{MIL}$, $L_{ann}$ and $L_{neg}$ to obtain $L_{CPR}$, Eq.~\ref{Eq:CPR basic loss};
\end{algorithmic}
}
\end{algorithm}

\textbf{CPR Loss.} Object-level MIL loss is introduced to endow CPRNet with the ability to find semantic points around each annotated point. Then to overcome the over-fitting problem of MIL when the data is insufficient, we further introduce the instance-level prior as supervision by designing annotation and negative loss. The objective function of CPRNet is a weighted summation of the three losses:
\vspace{-2pt}
\begin{equation}\small
\begin{aligned}
L_{CPR} = & L_{MIL} + \alpha_{ann} L_{ann} +\alpha_{neg} L_{neg},
\label{Eq:CPR basic loss}
\end{aligned}
\end{equation}
\vspace{-2pt}where $\alpha_{ann}=0.5$ and $\alpha_{neg}=3$ (by default in this paper). And $L_{MIL}$, $L_{ann}$; $L_{neg}$ are based on the focal loss~\cite{DBLP:conf/iccv/LinGGHD17}:
\vspace{-6pt}
\begin{equation}\small
\begin{aligned}
FL(S_{p}, c_{j}) = \sum\limits_{k_c=1}^{K_c} c_{j,k_c}& (1 - S_{p, k_c})^{\gamma} \log(S_{p, k_c}) + \\
&(1-c_{j,k_c}) S_{p, k_c}^{\gamma} \log(1-S_{p, k_c}),
\label{Eq:focal loss}
\end{aligned}
\end{equation}
\vspace{-2pt}

\noindent where $\gamma$ is set as 2 in Eq.~\ref{Eq:focal loss}
following the standard focal loss, and $S_p \in \mathbb{R}^{K_c}$ and $c_{j} \in \{0, 1\}^{K_c}$ are the predicted scores on all categories and the category label, respectively.

\textbf{Object-level MIL Loss.} To find the semantic points during refinement, we refer to WSOD ~\cite{DBLP:conf/cvpr/BilenV16} and design a MIL loss to enable the CPRNet to justify whether the points in $B_j$ are in the same category with $a_j$. 
Based on $B_j$, the feature vectors $\{\mathbf{F}_p|p\in B_j\}$ are extracted. As Eq.~\ref{Eq:S_{bag}} shows, for each $p \in B_j$, a classification branch $fc^{cls}$ is applied to obtain the logits $[\mathbf{O}^{cls}_{B_j}]_p$, which is then utilized as an input of an activation function $\sigma_1$ to obtain $[\mathbf{S}^{cls}_{B_j}]_p$. Besides, an instance selection branch $fc^{ins}$ is applied to $\mathbf{F}_p$ to obtain $[O^{ins}_{B_j}]_p$, which is then utilized as an input of an activation function $\sigma_2$ to obtain the selection score $[S^{ins}_{B_j}]_p$. The score $[S^{over}_{B_j}]_p$ is obtained by taking the element-wise product of $[S^{ins}_{B_j}]_p$ and $[S^{cls}_{B_j}]_p$.
\vspace{-2pt}
\begin{equation}\small
\begin{aligned}
&{[O^{cls}_{B_j}]}_p = fc^{cls}(F_p) \in \mathbb{R}^{K_c}, \quad
[O^{ins}_{B_j}]_p = fc^{ins}(F_p) \in \mathbb{R}^{K_c}; \\
&[S^{cls}_{B_j}]_p = [\sigma_1(O^{cls}_{B_j})]_p = 1/(1+e^{-[O^{cls}_{B_j}]_p})\in \mathbb{R}^{K_c}; \\
&[S^{ins}_{B_j}]_p = [\sigma_2(O^{ins}_{B_j})]_p = e^{O^{ins}_p}/\sum_{p' \in B_j} [e^{O^{ins}_{B_j}}]_{p'} \in \mathbb{R}^{K_c}; \\
&[S^{over}_{B_j}]_p = [S^{ins}_{B_j}]_p \cdot [S^{cls}_{B_j}]_p \in \mathbb{R}^{K_c},
\label{Eq:S_{bag}}
\end{aligned}
\end{equation}
\vspace{-8pt}

\noindent where $\sigma_2$ is a $softmax$ function. Different from MIL in WSOD, the $sigmoid$ activation function is applied for $\sigma_1$, due to its suitability for the binary task compared with the $softmax$ function. Furthermore, the $sigmoid$ activation function allows to perform of multi-label classification (for the overlapping area of multiple objects' neighborhoods) for points and is more compatible with focal loss.

The bag-level score $S_{B_j}$ is obtained by the summation of all points' scores in $B_j$ as shown in Eq.~\ref{Eq:S_{bag}2}. $S_{B_j}$ can be seen as the weighted summation of the classification score $[S^{cls}_{B_j}]_p$ of $p$ in $B_j$ by the corresponding selection score $[S^{ins}_{B_j}]_p$. 
\begin{equation}\small
\begin{aligned}
S_{B_j} & = \sum_{p \in B_j} [S^{over}_{B_j}]_p \in \mathbb{R}^{K_c}.
\label{Eq:S_{bag}2}
\end{aligned}
\end{equation}

Finally, the MIL loss is calculated using the focal loss on the predicted bag-level scores $S_{B_j}$ and the category label $c_j$ of $a_j$:
\vspace{-7pt}
\begin{equation}\small
\begin{aligned}
L_{MIL} & = \frac{1}{M}\sum\limits_{j=1}^{M}FL(S_{B_j}, c_j).
\label{Eq:L_{bag}}
\end{aligned}
\end{equation}
\vspace{-7pt}

\textbf{Annotation Loss.} Due to the lack of explicit positive samples for supervision in MIL, the network sometimes focuses on the points outside the instance region and mistakenly regards them as the foreground.
Therefore, we introduce the annotation loss $L_{ann}$, which gives the network accurate positive samples for supervision via annotated points, to guide MIL training. $L_{ann}$ can guarantee a high score of the annotated point and mitigate misclassification to some extent. 
Firstly, the classification score of $S_{a_j}(j \in {1, 2, ...M})$ of $a_j$ is calcluated as:
\vspace{-7pt}
\begin{equation}\small
\begin{aligned}
S_{{a_j}} &= \sigma_1(fc^{cls}(F_{a_j})) \in \mathbb{R}^{K_c}.
\label{Eq:S_{a_j}}
\end{aligned}
\vspace{-5pt}
\end{equation}
$L_{ann}$ is calculated with focal loss as:
\vspace{-8pt}
\begin{equation}\small
\begin{aligned}
L_{ann} &= \frac{1}{M}\sum\limits_{j=1}^{M} FL(S_{a_j}, c_j).
\label{Eq:SL_{a_j}}
\end{aligned}
\end{equation}
\vspace{-10pt}

\textbf{Negative Loss.} The conventional MIL adopts binary log loss, and it views the proposals belonging to other categories as negative samples. 
To lack of explicit supervision from samples in the background, the negative samples are not well suppressed during MIL training.
Therefore, based on $Neg_k$, the negative loss $L_{neg}$, the negative part of focal loss, is calculated as:
\vspace{-5pt}
\begin{equation}\small
\begin{split}
S_p &= \sigma_1(fc_{cls}(F_p)) \in \mathbb{R}^{K_c}; \\
L_{neg} &= \frac{1}{M}\sum\limits_{k_c=1}^{K_c} \sum\limits_{p\in Neg_{k_c}} c_{j,k_c} S_{p, k_c}^{\gamma} \log(1 - S_{p, k_c}).
\label{Eq:S_{Neg}}
\end{split}
\end{equation}
\vspace{-5pt}

\noindent where $\gamma$ is set as 2.

\begin{algorithm}[tb!]
{
\small
\caption{CPRInfer}
\label{Alg: CPRInfer}
\textbf{Input:} Annotated point $A$, Sampling center $\hat{A}$ and radius $\hat{R}$ of image $I$. \\
\textbf{Output:} Semantic points $B^+$ and their scores $S^+$ of each instance in $I$.\\
\textbf{Note:} $\mathbf{F}$ is extracted feature map of image $I$ according to Sec. \ref{sec:CPRNet architecture}; $\delta_1, \delta_2$ are thresholds. $K$ is the number of categories. $S_{p,k}$ is the predicted score on $k$-th category of point $p$. $k_c \in \{1, 2, ...K\}$ is the category label (not one-hot format).\\
\vspace{-10pt}
\begin{algorithmic}[1]

\STATE $B^+, S^+ \leftarrow \{\}, \{\}$;

\FOR{$a_j \in A$, $\hat{a}_j \in \hat{A}$, $\hat{r}_j \in \hat{R}$}
\STATE $B_j \leftarrow BS(a_j, \hat{a}_j, \hat{r}_j)$, Eq.~\ref{Eq:Bag_j};
\STATE $S_{a_j} \leftarrow \sigma_1 (fc^{cls}(\mathbf{F}_{a_j}; \hat{E})) \in \mathbb{R}^{K_c}$;
\STATE $B^+_j, S^+_j \leftarrow \{a_j\}, \{S_{a_j}\}$;
\FOR{$p \in B_j$}
\STATE $S_p \leftarrow \sigma_1 (fc^{cls}(\mathbf{F}_p; \hat{E})) \in \mathbb{R}^{K_c}$;
\STATE find $k_c \in \{1, 2, ..K_c\}$ s.t. $c_{jk_c} = 1$;
\STATE $s_p \leftarrow S_{p,k_c}$;
\IF{{$s_p > \delta_1$ and $s_p > \delta_2 * S_{a_j, k_c}$ and
\\ \ \ \ \ $k_c = \text{argmax}_{1 \leq k_c\leq K_c}\ S_{p, k_c}$ and
\\ \ \ \ \ $a_j = \text{argmin}_{a \in A}\ ||p-a||$\ \ \ }
}

\STATE $B^+_j, S^+_j \leftarrow B^+_j \cup \{ p \}, S^+_j \cup \{ s_p\}$;
\ENDIF
\ENDFOR
\STATE $B^+, S^+ \leftarrow B^+ \cup \{B^+_j \}, S^+ \cup \{S^+_j \}$;
\ENDFOR
\end{algorithmic}
}
\end{algorithm}

\begin{figure*}[tb!]
\begin{center}
    \begin{tabular}{ccc}
    \includegraphics[width=0.98\linewidth]{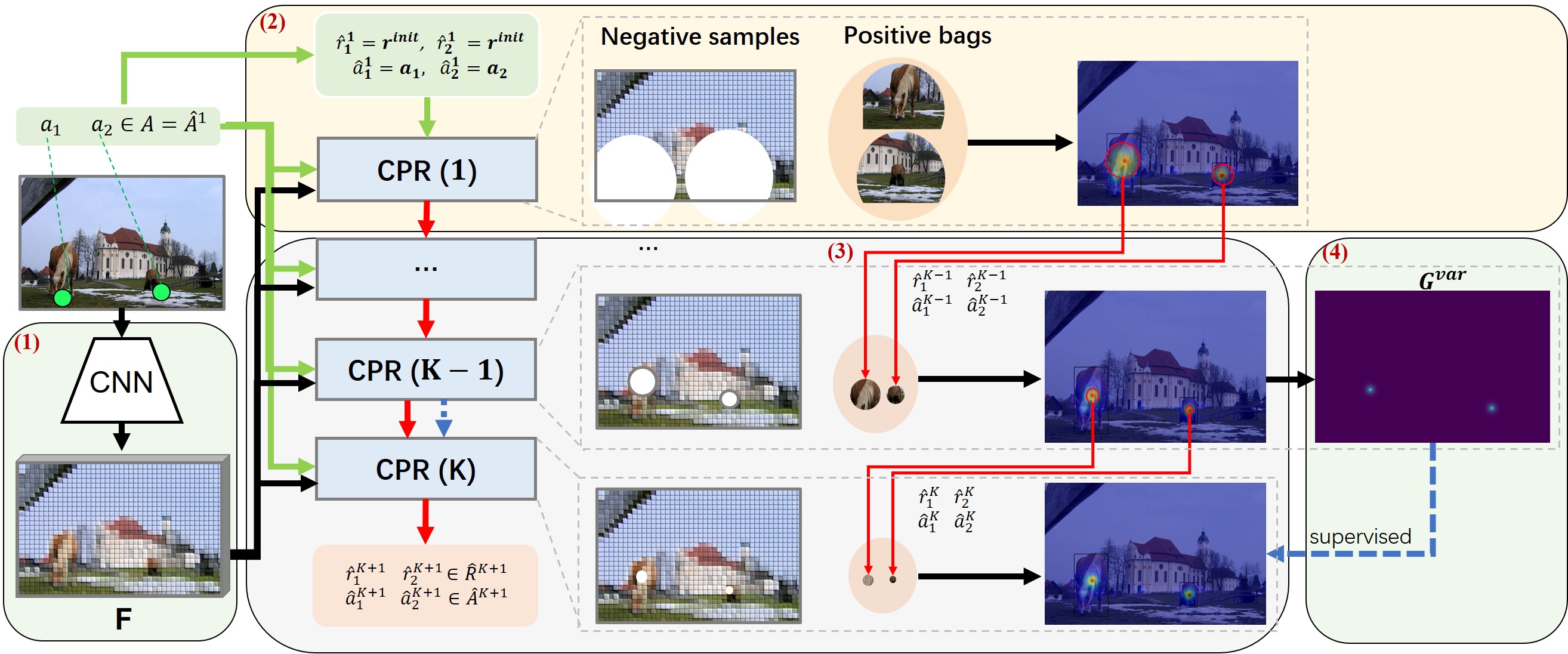}
    \end{tabular}
    \caption{The framework of CPR++. (1) For an image, a backbone is used to extract the shared feature map for all the CPR heads. (2) Given the annotation point as a center point and initial large radius for the sampling region, the positive points bag, and negative samples are constructed to train the first CPR head. The semantic points estimated by the CPR head are utilized to obtain the adaptive smaller sampling radius and the semantic center point by sampling region estimation module. (3) The dynamic sampling radius and semantic center point are utilized to format the new positive points bag and negative samples to train the next CPR head. Repeating these procedures $K$ times. (4) For the last ($K$-th) CPR head, variance regularization is conducted with the supervision of the variance map $G^{var}$ generated by the previous stage to reduce semantic variance further. It is worth mentioning that the initial annotation points are utilized in all heads to calculate annotation loss. CPR++ outputs the $\hat{A}^{K+1}$ as the final refined point to supervise the localizer.}
\label{fig:method CCPR}
\end{center} 
\vspace{-15pt}
\end{figure*}

\subsubsection{CPR Inference} \label{sec: CPRInfer}
As described in Algorithm ~\ref{Alg: CPRInfer}, the trained network $\hat{E}$ is utilized to obtain semantic points $B^+$ and their scores $S^+$.
Based on $B_j$, $[S^{cls}_{B_j}]_p$ predicted by $\hat{E}$ and the constraints (details given in following), points that are the same category (similar semantic) as the annotated point are selected, denoted as $B^+_j$, their scores denoted as $S^+_j$. 
To obtain $B^+_j$, three constraints (line 10 in Algorithm~\ref{Alg: CPRInfer}) are introduced. The Constraint I is to delete points with small classification scores. We filter out the point $p \in B_j$ whose $S_{p, k_j}$ is smaller than the threshold $\delta_1$ (set as 0.1 by default) or $\delta_2*S_{a_j, k_j}$, where $\delta_2$ is set as 0.5 by default and $k_j$ is category label (not one-hot format) of $j$-th object. 
Constraint II is to delete the points that are not classified correctly. Specifically, the correct classification means the classification score $S_{p, k_j}$ of point $p$ on the given annotated category $k_j$ is higher than the scores on other categories. 
Constraint III is to delete the points closer to another object in the same category. Since two adjacent objects of the same category may interfere with each other. With the three constraints, the remaining points $B^+_j$ and their scores $S^+_j$ are collected.

Finally, we weighted average $B^+_j$ with $S^+_j$ to obtain the semantic center point(the final refine point) as supervision to train localizer which will be specifically described in experiment section.

\vspace{-15pt}
\subsection{CPR++}\label{sec:cpr++}

\label{sec:CPR++}
\subsubsection{Motivation}

Without access to object scale information, CPR uses a pre-defined and fixed sampling region (radius $\hat{r}$) for all objects.
While this practice works for most objects, it may be limited in some cases.
For example, when using a very large sampling radius $\hat{r}$, many background points will be included in the points bags and then MIL might not be effective in suppressing them. Also, for adjacent objects, the large $\hat{r}$ makes the sampling area of different objects overlap, which might confuse the training and lead to localization mismatch as shown in Fig.~\ref{fig:motivation of ccpr} (c). 
In addition, a small $\hat{r}$ can alleviate the above two cases, but it limits the refining solution space for large objects and degrades the localization performance on them.
To further study the effect of the sampling region, we report the results of using different $\hat{r}$ in Fig.~\ref{fig:motivation of ccpr} (d) and Table~\ref{tab: different $R$ for CPR}. {We observe that as the value of $\hat{r}$ increases, the overall performance (mAP) on small objects decreases; the performance on large objects increases first and then decreases. The mAP drop of large objects is probably caused by the adjacency and background issues illustrated above.

The above analysis shows it is crucial to use a suitable sampling region for each object. Driven by this, we propose to estimate the sampling regions of each object. That is, we adaptively set the radius and centre of each object during training and {inference}, yielding CPR++.
}
It contains three parts: 1) Sampling Region Estimation: a spatial variance and semantic center point of each instance are estimated from the semantic points predicted by CPR inference. 
2) Progressive Point Refinement: the center point and spatial variance are used as the new sampling center and radius to calculate a new CPR loss and predict new semantic points. Using a cascade structure, we repeat this procedure and progressively refine the semantic points. Moreover, a variance regularization is introduced to boost the refinement. 3) CPR++ Inference: the semantic center points in the final stage are used as the final refined point to train localizer.

\subsubsection{Sampling Region Estimation}\label{sec: variance estimation}



The sampling region of each object is defined by the radius and centre. In this section, we introduce a heuristic method to estimate the radius and centre for each object based on its semantic points $B^+_j$ given by CPR.
As formulated in Algorithm~\ref{Alg: update_sample_region} and shown in right of Fig.~\ref{fig:method CPR}, in order to generate a dynamic and adaptive sampling region for objects with different scales, 
the semantic center point $\hat{a}$, calculating by averaging semantic points $B^+_j$ with their scores $S^+_j$, is utilized as new sampling centre
and the spatial variance estimated from Eq.~\ref{Eq:S_{variance estimation}}, which can approximate object's  scale information to some extend, is utilized as new sampling radius. 
To be specific, a minimum bounding box $box_j=([box_j]_{x1}, [box_j]_{y1}, [box_j]_{x2}, [box_j]_{y2})$ of the semantic points $B^+_j$ is generated and then $box_j$'s size (the root of the area), $\hat{r}_j$ is calculated to measure the spatial variance of semantic points. 
The idea behind it comes from the fact: the point annotation of large objects usually have large semantic variance because there are more optional points to annotate for larger object. This means larger object contains more semantic points. Therefore we utilize the spatial variance of the semantic points to design a scale-adaptive sampling radius. 

\vspace{-15pt}
\begin{equation}\small
\begin{split}
[box_j]_{x1}, [box_j]_{y1} = \min_{(p_x, p_y \in B^+_j)} p_x, \min_{(p_x, p_y \in B^+_j)} p_y;\\
[box_j]_{x2}, [box_j]_{y2} = \max_{(p_x, p_y \in B^+_j)} p_x, \max_{(p_x, p_y \in B^+_j)} p_y \\
\hat{r}_j = \sqrt{([box_j]_{x2} - [box_j]_{x1}) * ([box_j]_{y2} - [box_j]_{y1})}.
\label{Eq:S_{variance estimation}}
\end{split}
\end{equation}

The $\hat{r}_j$ and $\hat{a}$ are utilized as the sampling radius and centre for a new CPR in the next stage to achieve adaptive and dynamic sampling for objects in different scales.



\begin{algorithm}[tb!]
{
\small
\caption{Sampling Region Estimation}
\label{Alg: update_sample_region}
\textbf{Input:} semantic points $B^+$ and their scores $S^+$\\
\textbf{Output:} Refined points $\hat{A}$ and sampling radius $\hat{R}$.\\
\vspace{-10pt}
\begin{algorithmic}[1]
\STATE $\hat{A}, \hat{R} \leftarrow \{\}, \{\}$;
\FOR{$B^+_j \in B^+$}
\STATE Calculate variance $\hat{r}_j$ with $B^+_j$, Eq ~\ref{Eq:S_{variance estimation}};
\STATE $\hat{a}_j \leftarrow (\sum_{p\in B^+_j} S^+_p * p) / (\sum_{p\in B^+_j} S^+_p)$;
\STATE $\hat{A}, \hat{R}^{k_s + 1} \leftarrow \hat{A} \cup \{\hat{a}_j\}, \hat{R} \cup \{\hat{r}_j\}$;
\ENDFOR

\end{algorithmic}
}

\end{algorithm}

\subsubsection{Progressive Point Refinement}

\textbf{I. Progressive Refinement.} Sampling region estimation enables us to refine the semantic points of objects progressively. Specifically, the sampling region estimation updates the sampling bag of each object. Based on this, we conduct a new round of CPR. By iteratively repeating the sampling region estimation and CPR, we can compute more accurate semantic points for objects.

The above progressive refinement is illustrated in Fig.~\ref{fig:method CCPR}, where CPR++ sequentially conducts $K$ CPR procedures.
The first stage is the standard CPR: given the initial annotated point $a_j \in A$, a large and fixed radius (super parameter) $r^{init}$ is chosen for the initial sampling region ($\hat{a}^{1}_j = a_j$ and $\hat{r}^{1}_j = r^{init}$). Then, the annotated points, sampled positive bags, and negative samples are fed to the CPR module to calculate CPR loss $L^{1}_{CPR}$.
For the next $k_s$-th ($k_s>1$) stage, the sampling center $\hat{a}^{k_s}_j$ and radius $\hat{r}^{k_s}_j$ are set as the $\hat{a}'$ and $\hat{r}'$, which are calculated by the sampling region estimation based on the semantic points of the ($k_s-1$)-th stage.
With a new sampling region, the loss $L^{k_s}_{CPR}$ is calculated and back-propagated to update the network. 
As the CPR procedure is condcuted multiple times, the semantic points for each subject will gradually become more precise.
We would like to mention that the progressive refinement based on dynamic sampling region has multiple implementations. We have tried some other designs but only the method given in this section works well. The details have been discussed in Sec.~\ref{sec: cascade designing}.

\noindent\textbf{II. Variance Regularization.} To further help the progressive refinement, we introduce a scale variance regularization.
\label{sec:variance-loss}
%
The objective of our optimization is to minimize the semantic variance of annotated points by adjusting their position. One of the necessary conditions for the objective is that the network trained with $\hat{A}$ as the supervision should predict a high score only in points $\hat{A}$, while the score is close to 0 in other points, which means that there is neither false alarm nor object omission.
To further reduce the semantic variance, we propose variance loss to promise the final refined points $\hat{A}$ to stratify the condition.
The extracted feature map $\mathbf{F}$ is passed through a convolution layer $Conv^{var}$, followed by a sigmoid layer $\sigma_2$, to obtain $S^{var}$: 
\begin{equation}
S^{var} = \sigma_2(Conv^{var} (\mathbf{F})) 
\label{Eq:var score}
\end{equation}
To achieve the condition mentioned before, the supervision $HG^{var} \in \mathbb{R}^{}$ of $S^{var}$ is defined as:
\begin{equation}
{HG}^{var}_{p,k_c} =\left\{
\begin{aligned}
1 & , & if\ p \in \hat{A}\ and\ c_{pk_{c}}=1, \\
0 & , & otherwise. 
\end{aligned}
\right.
\label{Eq:var supervision hard}
\end{equation}
where $HG^{var}_{p,k_c}$ represents the supervision of point $p$ on $k_c$-th category.
However, this kind of hard supervision (0 or 1) is difficult to optimize. So a Gaussian-like function is used for approximation as defined in Eq. ~\ref{Eq:var supervision gaussian}. First for each point $\hat{a_j} \in \hat{A}$, a Gaussian map $G^{*}_j$ is generated. And then, we aggregate all Gaussian maps for each category by element-wise max function to obtain the soft supervision $G^{var}$. 
\begin{equation}
\centering
\begin{split}
G^{*}_{p,j} = {exp({-\sqrt{(p_x - \hat{a}_{j,x})^2 + (p_y - \hat{a}_{j, y})^2} / \sigma})}  \in \mathbb{R}. \\
G^{var}_{p,k_c} = \max_{\hat{a}_j\in \hat{A}_{k_c}}G^{*}_{p, j} \in \mathbb{R}.
\end{split}
\label{Eq:var supervision gaussian}
\end{equation}
where $\hat{A}_{k_c}$ is set of $k_c$-th category points in $\hat{A}$.

Finally, the cross entropy function is applied to calculate the loss between $G^{var}$ and $S^{var}$:
\begin{equation}
L_{var}(\hat{A}) = -\sum_{p,k_c} G^{var}_{p,k_c} log S^{var}_{p,k_c} + (1 - G^{var}_{p,k_c}) log (1-S^{var}_{p,k_c})
\label{Eq:var loss}
\end{equation}
where $S^{var}_{p,k_c}$ is var score of $k_c$-th category of point $p$.

\noindent\textbf{III. The overall loss function.} Combining $L^{k_s}_{CPR}$ of all stages and $L_{var}(\hat{A}^{K}$), the $L_{CPR++}$ is obtained as:
\begin{equation}
L_{CPR++} = \sum_{k_s=1}^{K} L^{k_s}_{CPR} + L_{var}(\hat{A}^{K}).
\label{Eq:CPR++ loss}
\end{equation}

The variance loss is less tolerant to noise due to the strict formulation compared with MIL. It can only be utilized in stages whose points have small variances, otherwise, it will lead to poor performance. In CPR++, the variance loss is only applied to the $K$-th stage (final stage), with the supervision coming from $(K-1)$-th stage.


\begin{algorithm}[tb!]
{
\small
\caption{CPR ++}
\label{Alg: CPR++}
\textbf{Input:} Training set $D_{train}$, network $E$. \\
\textbf{Output:} Refined points $\hat{A}^{D_{train}}$. \\
\textbf{Note:} $A$ and $C$ are 2D coordinates and category label of annotated points in image $I$ respectively. $r^{init}$ is initial sampling radius (super parameter). $K$ is number of stages. when $K=1$ and removing $L_{var}$, CPR++ degenerates into origin CPR.\\
\vspace{-10pt}
\begin{algorithmic}[1]
\STATE $R^{init} \leftarrow \{r^{init}, r^{init}, ...\}$;
\STATE \textit{// training step}
\STATE $L_{CPR++}^{D_{train}} \leftarrow 0$;
\FOR{$(I, A, C) \in D_{train}$}
\STATE Extract feature map $\mathbf{F}$ of $I$ with $E$;
\STATE $\hat{A}^{1}, \hat{R}^{1} \leftarrow A, R^{init}$;
\FOR{$k_s = 1, 2...,K$}
\STATE $L_{CPR} \leftarrow CPRLoss(A, \hat{A}^{k_s}, \hat{R}^{k_s};\mathbf{F})$, Algorithm~\ref{Alg: CPRLoss};
\STATE $L_{CPR++}^{D_{train}} \leftarrow L_{CPR++}^{D_{train}} + L_{CPR}$;
\IF{$k_s = K_S$ and $K_S \neq 1$}
\STATE $L_{CPR++}^{D_{train}} \leftarrow L_{CPR++}^{D_{train}} + L_{var}$;
\ENDIF
\STATE $B^+, S^+ \leftarrow CPRInfer(A, \hat{A}^{k_s}, \hat{R}^{k_s};\mathbf{F})$, Algorithm~\ref{Alg: CPRInfer};
\STATE $\hat{A}^{k_s + 1}, \hat{R}^{k_s + 1} \leftarrow \text{sample}\_\text{region}\_\text{estimation}(B^+, S^+)$, Algorithm~\ref{Alg: update_sample_region};
\ENDFOR
\ENDFOR
\STATE Train $E$ by minimizing $L_{CPR++}^{D_{train}}$ to obtain $\hat{E}$.
\STATE \textit{// inference step}
\STATE $\hat{A}^{D_{train}} \leftarrow \{\}$;
\FOR{$(I, A, C) \in D_{train}$}
\STATE Extract feature map $\mathbf{F}$ of $I$ with $\hat{E}$;
\STATE $\hat{A}^{1}, \hat{R}^{1} \leftarrow A, R^{init}$;
\FOR{$k_s = 1, 2...,K$}
\STATE $B^+, S^+ \leftarrow CPRInfer(A, \hat{A}^{k_s},\hat{R}^{k_s};\mathbf{F})$
\STATE $\hat{A}^{k_s + 1}, \hat{R}^{k_s + 1} \leftarrow \text{sample\_region\_estimation}(B^+, S^+)$, Algorithm~\ref{Alg: update_sample_region};
\ENDFOR
\STATE $\hat{A}^{D_{train}} \leftarrow \hat{A}^{D_{train}} \cup \{\hat{A}^{K + 1}\}$;
\ENDFOR
\end{algorithmic}
}
\end{algorithm}

\subsubsection{CPR++ Inference}

\begin{figure*}[htbp]
\begin{center}
    \includegraphics[width=0.98\linewidth]{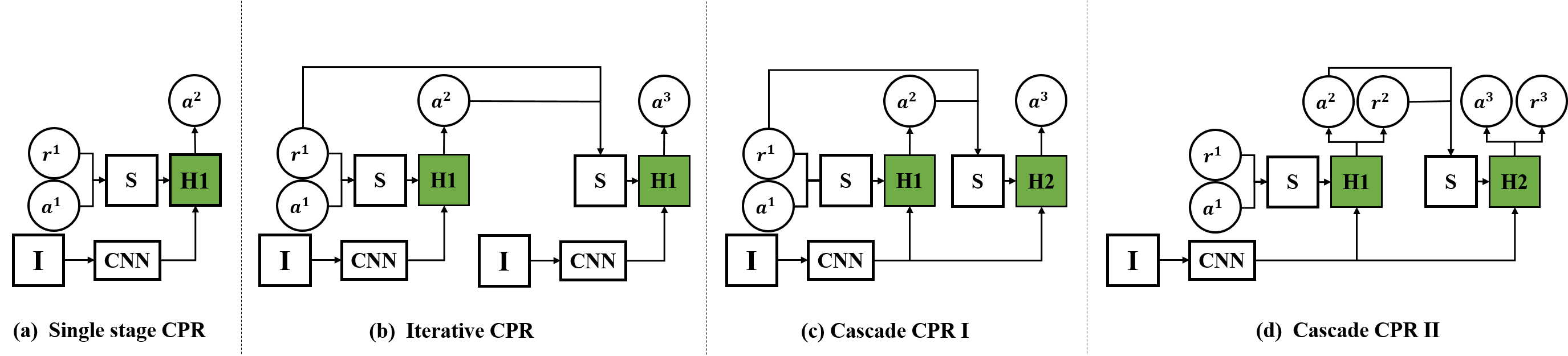}
\caption{Illustrations of different cascade modes. In each subfigure, ``I" is the input image, ``CNN" denotes the backbone convolutions, ``H" is the CPR network head, ``S" denotes the point sampling, ``a" is annotation, and ``r" is the sample radius. For simplicity, we only show two stages of cascade modes.}
\label{fig:cascade mode}
\end{center}
\vspace{-20pt}
\end{figure*}

The whole training and inference procedure of CPR++ are described in Algorithm~\ref{Alg: CPR++}.
The trained network of CPR++, $\hat{E}$ is utilized to extract feature map $\mathbf{F}$ of each image $I$. And then, $R^{init}$ and the annotated point $A$ are chosen as the initial sampling radius $\hat{R}^1$ and centre $\hat{A}^1$, which is the same as the training procedure. For $k_s$-th stage, $CPRInfer$ module is performed with $A$, $\hat{A}^{k_s}$ and $\hat{R^{k_s}}$ based on $\mathbf{F}$. And then $update\_sample\_region$ (Algorithm~\ref{Alg: update_sample_region}) is utilized to obtain new sampling radius $\hat{R}^{k_s+1}$ and centers $\hat{A^{k_s+1}}$.
With all these stages, CPR ++ progressively refines the sampling centres. The sampling centre of the final stage, $\hat{A}^{K+1}$, is chosen as the final refined point, which is used as the supervision to train P2PNet~\cite{Song_2021_ICCV}.
P2PNet is the SOTA baseline for the POL task and will be specifically described in the experiment section.

\begin{figure*}[tb!]
  \hfill
  \begin{subfigure}{1.0\linewidth}
    \centering
    \includegraphics[width=0.99\linewidth]{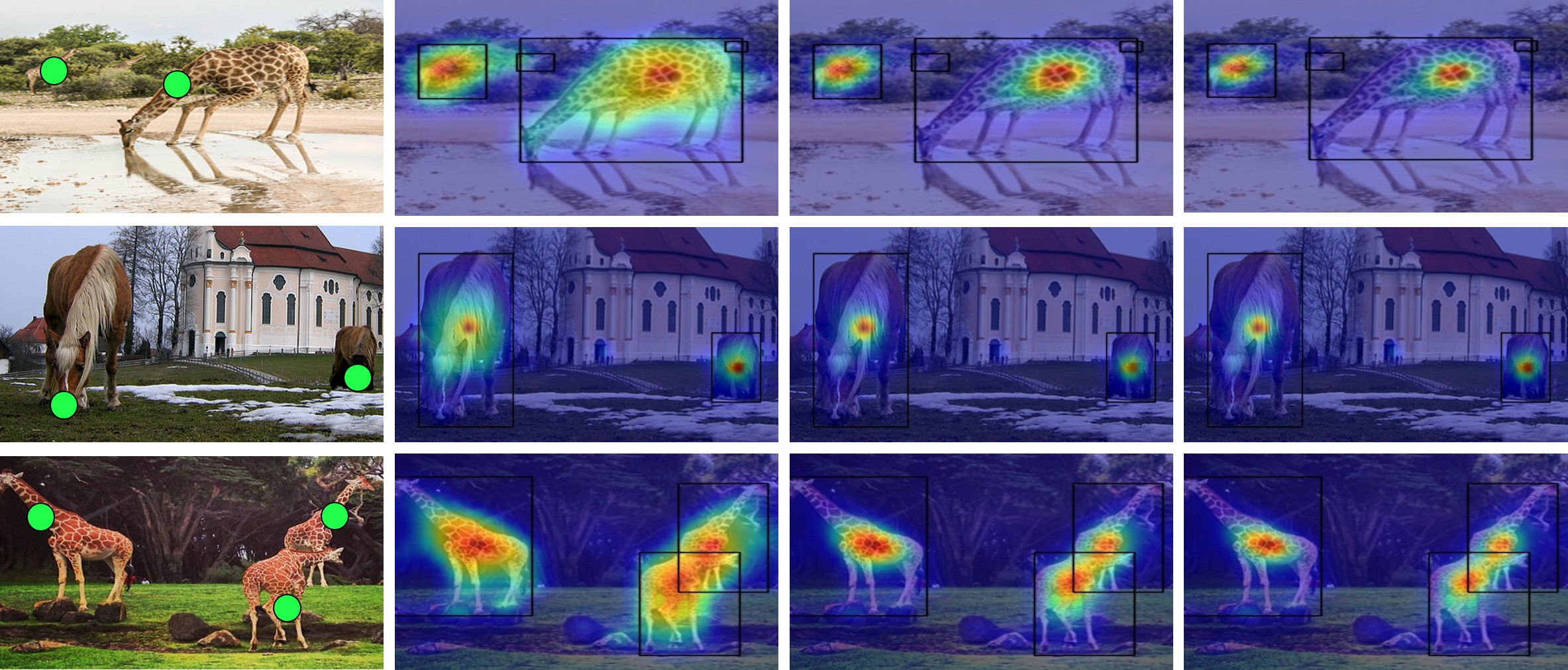}
  \end{subfigure}
  \caption{The visualization of CPR++. The right three columns are heatmaps indicating the semantic variance of the first, second and last stages of CPR++. We can see that the semantic variance gradually reduces in the cascade structure. \wj{The black bounding box is box-level ground-truth for better view}.}
  \vspace{-10pt}
\label{fig:visilization of CCPR head map}
\end{figure*}

With the progressive refinement structure and variance loss constraint mentioned above, the CPR++ can effectively mitigate semantic variance as shown in Fig.~\ref{fig:visilization of CCPR head map} and thus improve the performance of the localizer.


\subsubsection{Cascade Structure} \label{sec: cascade designing}

In this section, we discuss the various implementations to achieve progressive sampling and show that the strategy (\emph{Cascade CPR II}) is more suitable.

Let us revisit the sampling problem in CPR. Although it provides a way to optimize annotated points by point sampling and MIL, it struggles to show superiority in adapting to multi-scale objects.
The fixed sampling region limits the refining process in CPR. To be specified, a small sampling radius $R$ is detrimental for large objects because of the difficulty in obtaining global optima.
Table~\ref{tab: different $R$ for CPR} demonstrates that the performance of large objects improves when the sampling radius $R$ becomes larger.
Conversely, the performance of small objects decreases with increasing $R$. We argue that the specific reasons for this are two-fold. First, more noise (background point) is introduced in the point bags during the MIL. Second, the interference of adjacent objects is enhanced due to more overlap regions.


Therefore, the critical issue is whether it is possible to expand the sampling region while ensuring fewer noise (background points and objects overlap) introduced.
There are two ways to realize such a goal. The first (\textit{Iterative CPR}, \textit{Cascade CPR I}) is iteratively adjusting the center point to expand the reachable sampling region while keeping a small radius to limit the introduced noise.
Another implementation method  (\textit{Cascade CPR II}) is initial with a large sampling radius, and then finding an adaptive smaller radius and center point for each object by \textit{sampling region estimation}. As repeating this process, we gradually narrow down the sampling region and find more precise semantic points.


\textbf{Iterative CPR.}
The most direct way to obtain the optimal global solution is iteratively using CPR for point annotation refinement to achieve the coarse-to-fine purpose. As shown in Fig.~\ref{fig:cascade mode} (b), with the first CPR and original point annotation, the refined point is obtained as the pseudo point annotation. Then, the pseudo point annotation is viewed as the input to train the next CPR. And so on, the final refined point is obtained with these iterative independent CPRs.

\textbf{Cascade CPR I.}
\textit{Iterative CPR} is not trained end-to-end. Therefore, the structure of multiple independent CPR leads to discontinuous optimization, which means loss formed by the later sampling range cannot be updated for the previous networks.
To solve this issue, we design an end-to-end structure called \emph{cascade CPR I}, which constructs a multi-head CPR structure sharing same backbone and implements iterative refinement in a cascade form.
As shown in Fig.~\ref{fig:cascade mode} (c), the multi-head CPR allows each branch to have a different sampling range, while the cascade approach drives the model to achieve progressive refinement. The end-to-end format effectively avoids forgetting previously learned positive and negative sample ranges during training and ensures the continuity of optimization.

\textbf{Cascade CPR II.}
Although the two approaches mentioned above extend the sampling range by iteratively updating the sampling center point, their fixed sampling radius remains sub-optimal for solving multi-scale issues.
In addition, \textit{cascade CPR I} updates the sampling center point to expand the sampling range while keeping $R$ small and unchanged, which inevitably leads to some points in the sampling region of a head being outside the sampling region of the previous head. Because points outside sampling region are treated as negative samples, that means these points on different CPR heads may be given different supervision, confusing network learning.
Therefore, we present \textit{cascade CPR II}, which obtains the dynamic sampling region by updating both sample center point and radius, as shown in  Fig.~\ref{fig:cascade mode}(d). 
It takes a large radius and the annotated point as center point to obtain the initial sampling region and then finds an adaptive smaller radius and center point for each object by \textit{sampling region estimation}. Repeating this process, it gradually narrows down the sampling region and finds more precise semantic points.
Because the new sampling radius is calculated based on the range of the previous CPR head predicting semantic points, which all fall within the range of the previous sampling region. 
Therefore, it ensures that the negative samples in the previous head are essentially negative in the next as well, with no conflict of supervision, which facilitates the optimization of the network.
The experiment results of these implementations are given in Table~\ref{tab:cascade mode} (first four lines). Only the Cascade CPR II improves the performance of large objects while maintaining the performance of the small objects.





\begin{figure}[tb!]
  \hfill
  \begin{subfigure}{1.0\linewidth}
    \centering
    \includegraphics[width=0.99\linewidth]{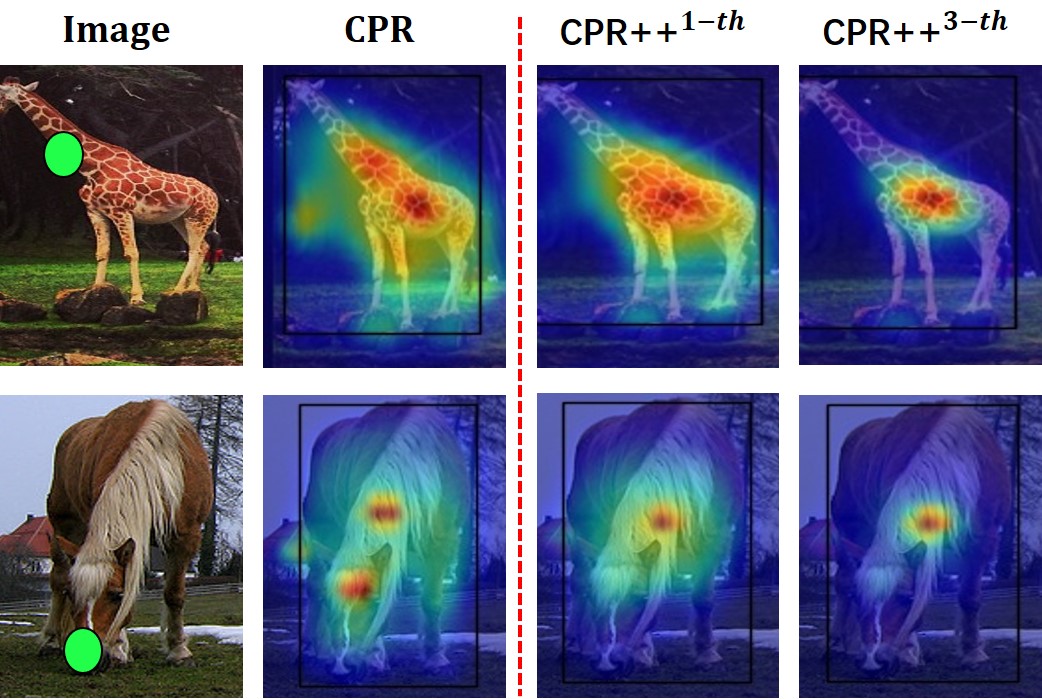}
  \end{subfigure}
  \vspace{-5pt}
  \caption{The comparison between CPR and CPR++. The CPR++$^{k-th}$ represents $k$-$th$ stage of CPR++.  CPR++ is more insensitive to the annotated point. (details in Sec.~\ref{sec: visualization analysis})  \wj{The black bounding box is box-level ground-truth for better view}}
  \vspace{-12pt}
\label{fig:visilization of CCPR vs CPR head map}
\end{figure}

\begin{figure}[tb!]
\hfill
\begin{subfigure}{1.0\linewidth}
\centering
\includegraphics[width=0.99\linewidth]{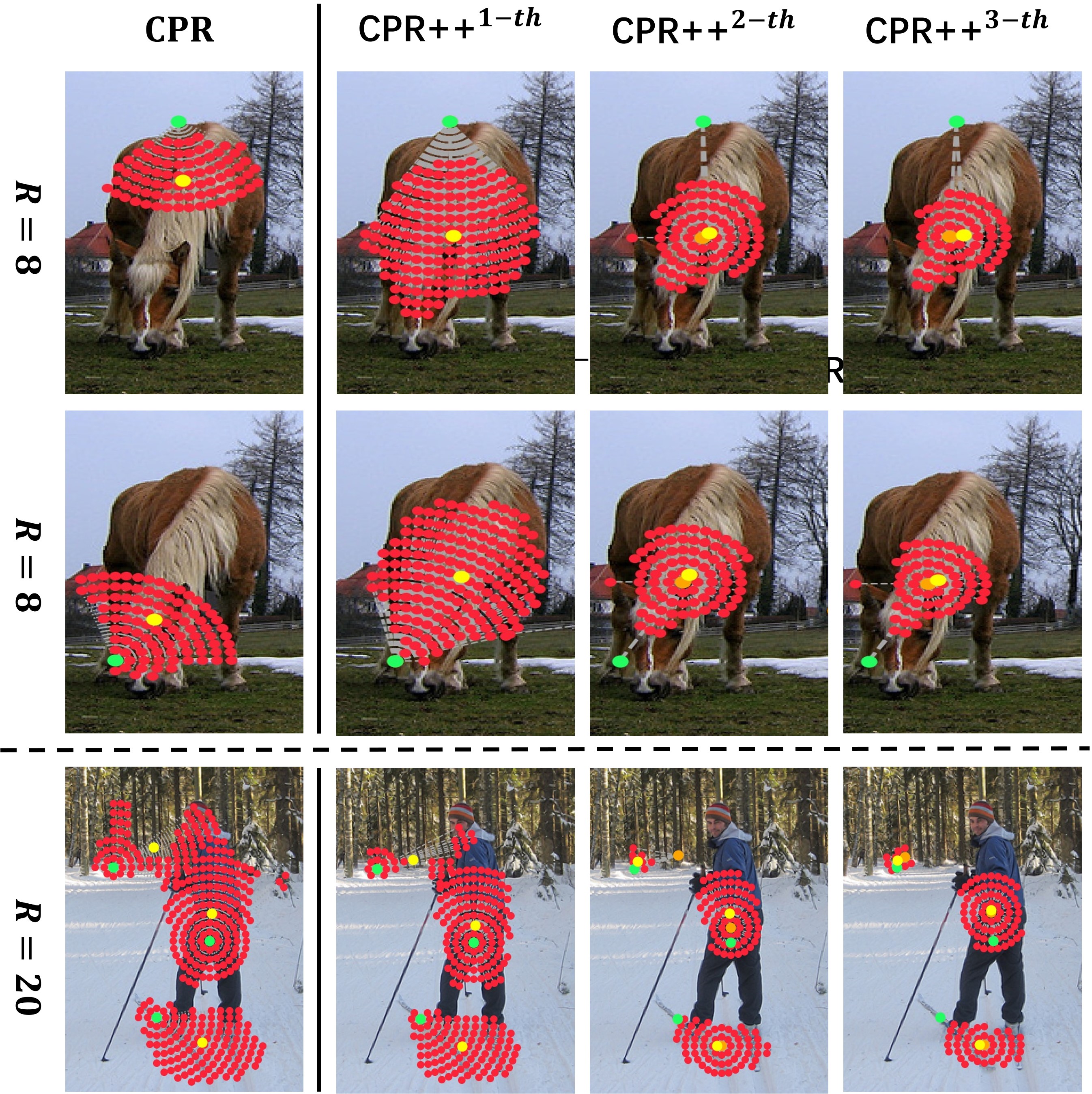}
\end{subfigure}
\vspace{-5pt}
\caption{The comparison of semantic points (red points) and refined points (yellow points) between CPR and different stages of CPR++, the green points is annotated point.
}
\vspace{-12pt}
\label{fig:visilization of CPR-CCPR}
\end{figure}

\subsubsection{Visualization Analysis} \label{sec: visualization analysis}

In this section, we give a visual analysis of CPR and CPR++.

\textbf{CPR++ progressively reduces the semantic variance.} Fig.~\ref{fig:visilization of CCPR head map} shows the spatial variance of semantic points in the first stage of CPR++ has a strong correlation with the object's scale. And then, the variance gradually reduces stage by stage. As the last row in Fig.~\ref{fig:visilization of CCPR head map} shown, the two giraffes on the right interfere with each other at first. While in the later stage, they are separated with the concentration of corresponding areas. This shows the advantages of CPR++.

\textbf{CPR++ is insensitive to the annotated points.} As shown in Fig.~\ref{fig:visilization of CCPR vs CPR head map}, on the heatmap of CPR++, there is only a single local-max point that is almost a semantic center point, while there is an additional local-max point near the annotated point on the heatmap of CPR. This means CPR++ is more insensitive to the annotated points. In addition, the sampling region and training loss in the first stage of CPR++ are the same as the original CPR. This also shows that the later stages have a positive impact on the first stage of CPR++. As shown in Fig.~\ref{fig:visilization of CPR-CCPR} (the first two rows), with different initial annotated points (green points) for the same cattle, CPR++ almost finds the same point as the refined point (yellow point), which shows CPR++ is more insensitive to the initial annotated point to some extent in another aspect. 


\textbf{CPR++ finds a more global solution.} As shown in Fig.~\ref{fig:visilization of CPR-CCPR} (the first two rows), when the radius is small in CPR, the sampling region is local. Therefore the refined point is in the respective local solution space of each annotation point (The back and head of the cattle). But in CPR++, the sampling radius dynamically covers the object's scale, so the final refined points move to the same region, which is the global solution of the semantic center.

\textbf{CPR++ can resist the influence of adjacent objects to some extent.} As shown in Fig.~\ref{fig:visilization of CPR-CCPR} (last row), when the radius is large in CPR, the sampling regions of two close objects belonging to the same category will overlap each other. Therefore the refined point will be towed to another object's region in CPR, especially for small objects. However, in CPR++, the drift will be corrected with the cascade mode and dynamic sampling radius.

\vspace{-5pt}
\section{Experiment}

\subsection{Experimental Settings}
\subsubsection{Datasets and Evaluation}
\textbf{Datasets.} For experimental comparisons, four publicly available datasets are used for point supervised localization task: COCO~\cite{lin2014microsoft}, DOTA-v1.0~\cite{DBLP:conf/cvpr/XiaBDZBLDPZ18}, Pascal VOC~\cite{DBLP:VOC} and SeaPerson. 
\textbf{COCO} is MSCOCO 2017, and it has 118k training and 5k validation images with 80 common categories. Since the ground-truth on the test set is not released, we train our model on the training set and evaluate it on the validation set. 
\textbf{DOTA}(v1.0) provides 2,806 images with 15 object categories. We utilize the training set for the training and the validation set for evaluation.
\textbf{Pascal VOC} uses VOC2012 train and VOC2007 trainval (10728
images) for training, and validates on VOC2012 val (5823 images). 
\textbf{SeaPerson}\footnote{SeaPerson is a low-resolution tiny person dataset.
} is a dataset for tiny person detection collected through a UAV camera at the seaside. The dataset contains 12,032 images and 619,627 annotated persons with low resolution. The images in the SeaPerson are randomly selected as training, validation, and test sets with the proportion of 10:1:10. 

\textbf{Evaluation.} Similar to WSOD, a point-box distance, calculated between the point and box, is for experimental evaluation. Specifically, the distance $d$ between point $a=(x, y)$ and  bounding box $b=(x^c, y^c, w, h)$ is defined as:
\begin{equation}\small
d(a, b)={\sqrt{\left(\frac{x-x^c}{w}\right)^2+\left(\frac{y-y^c}{h}\right)^2}}.
\label{Eq:distance-base}
\end{equation}
where $(x^c, y^c)$, $w$, $h$ are the center point, width, and height of the bounding box, respectively. 
The distance $d$ is used as the matching criterion for POL performance. 
A point and the object's bounding box are matched if the distance $d$ is smaller than a predefined threshold $\tau$ (set as 0.5/1.0/2.0, respectively, meaning that as long as the point falls within an unmatched half-center region/whole region/double region ground-truth box, the point successfully matches the ground-truth box.). If a bounding box has multiple matched points, the point with the highest score is chosen. While a point has multiple matched objects, the object with the smallest point-box distance is selected. A true positive (TP) is counted if a point matches an object. Otherwise, a false positive (FP) is counted. Neither TP nor FP will be counted if a point matches an object that is annotated as ignore, which follows the evaluation criteria of pedestrian detection~\cite{zhang2017citypersons} and TinyPerson benchmark~\cite{Yu2020ScaleMF}.
We adopt Average Precision (AP) as the metric. 
${\rm mAP}_{\tau}$ represents Average Precision under threshold $\tau$ over all classes.
Because the task of coarse point annotations pays more attention to whether the objects can be found, the ${\rm mAP}_{1.0}$ is chosen as the main evaluation metric in the following experiments unless otherwise specified. To measure the performance of objects in different scales, ${\rm mAP}^{small}$(${\rm mAP}^{s}$), ${\rm mAP}^{mediu}$(${\rm mAP}^{m}$) and ${\rm mAP}^{large}$(${\rm mAP}^{l}$) (under $\tau$=1.0) are also included to evaluate the performance of objects whose areas are in [0, $32^2$], [$32^2$, $96^2$] and [$96^2$, $+\infty$], respectively.

\vspace{-6pt}
\subsubsection{SeaPerson Details}
\begin{figure}[tb!]
\begin{center}
    \begin{tabular}{lcc}
     \includegraphics[width=0.9\linewidth]{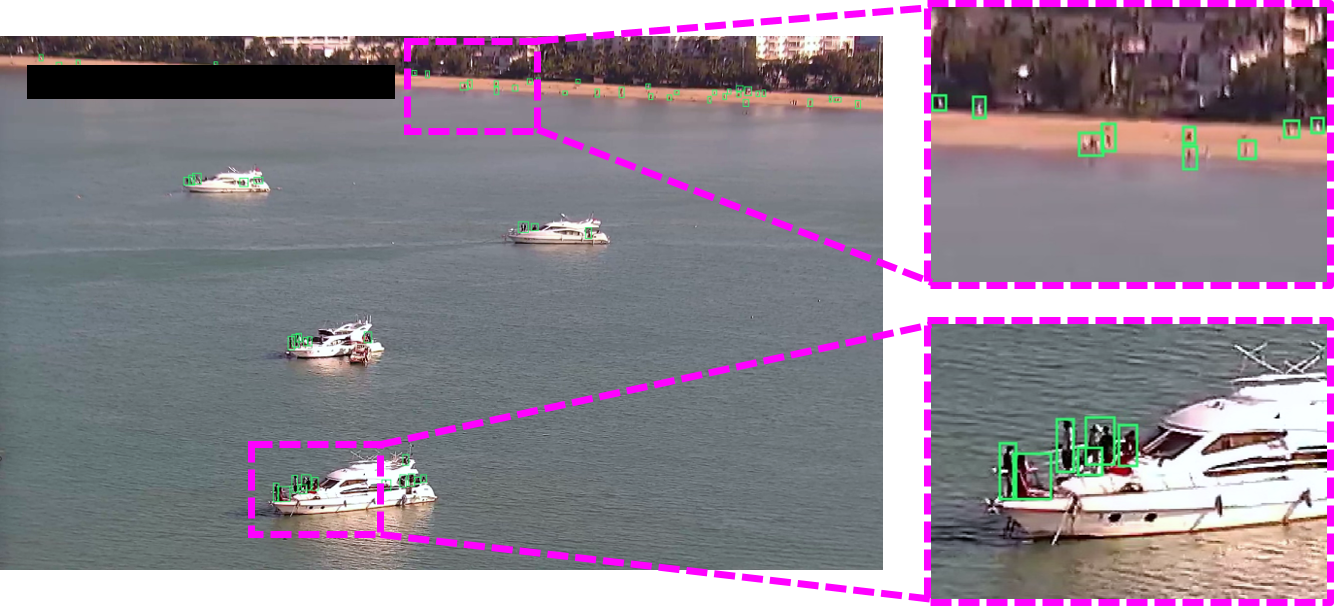}
    \end{tabular}
\setlength{\belowcaptionskip}{-0.5cm}
\vspace{-5pt}
\caption{Examples of SeaPerson. The resolution of image is very low and the objects in the dataset are extremely small.}
\label{fig: example of seaperson}
\end{center}
\vspace{-10pt}
\end{figure}

SeaPerson is built for tiny-person localization, which can help quick maritime rescue, beach safety inspection and so on.
The resolution of images is mainly 1920 $\times$ 1080, and the person size is extremely low (about 22.6 pixels). Therefore, there is no privacy-sensitive information in SeaPerson.
\textbf{1) Dataset Collection.} 
SeaPerson is collected as :
\romannumeral1) Videos are recorded in various seaside scenes by an RGB camera on an Unmanned Aerial Vehicle.
\romannumeral2) We sample an image of every 50 frames from the video and remove images with high homogeneity.
\romannumeral3) We annotated all persons in all sampled images with bounding boxes.
\romannumeral4) Following the rules of coarse point annotation in Sec.~\ref{sec3:Coarse Point Annotation}, coarse point annotation is obtained on SeaPerson for POL task.
\textbf{2) Dataset Splitting.} We randomly split the dataset into three subsets (training set, valid set and test set), while images from the same video sequence cannot be separated into different subsets. 
As shown in Table~\ref{tab: volumn of datasets}, the ratio of images' number in the training set, valid set and test set is about 10:1:10. 
\textbf{3) Dataset Properties.} Our proposed SeaPerson is similar to TinyPerson, while the volume of SeaPerson is about 7 times that of TinyPerson. The absolute size and relative size of objects are very small as shown in Table~\ref{tab: size of datasets} and Fig.~\ref{fig: example of seaperson}. In such scenario, we only care about the position of the object rather than the size of the object, which makes it very suitable for POL task. In addition, SeaPerson can also be used as a dataset for tiny object detection due to the bounding box annotation.

\begin{table}
\begin{center}
\resizebox{0.48\textwidth}{!}{
\begin{tabular}{l|c|c|c|c}
\specialrule{0.13em}{0pt}{0pt}
 & \multicolumn{2}{|c|}{SeaPerson} & \multicolumn{2}{|c}{TinyPerson} \\
\cline{2-5}
 & \#images & \#annotations & \#images & \#annotations \\
\hline\hline
train set  & 5711 & 262063 & 794  & 42197 \\
valid set  &  568 &  42399 & 816  & 30454 \\
test set   & 5753 & 315165 & -    & -\\
sum        & 12032& 619627 & 1610 & 72651 \\
\specialrule{0.13em}{0pt}{0pt}
\end{tabular}
}
\end{center}
\vspace{-4mm}
\caption{Statistic information of SeaPerson and TinyPerson.}
\vspace{-8pt}
\label{tab: volumn of datasets}
\end{table}

\begin{table}
\begin{center}
\resizebox{0.48\textwidth}{!}{
\begin{tabular}{l|c|c|c}
\specialrule{0.13em}{0pt}{0pt}
Dataset & absolute size & relative size & aspect ratio \\
\hline\hline
COCO       & 99.5 $\pm$ 107.5& 0.170 $\pm$ 0.203 & 1.213 $\pm$ 1.337\\
TinyPerson & 17.0 $\pm$ 16.9 & 0.011 $\pm$ 0.010 & 0.690 $\pm$ 0.422\\
SeaPerson  & 22.6 $\pm$ 10.8 & 0.016 $\pm$ 0.007 & 0.723 $\pm$ 0.424\\
\specialrule{0.13em}{0pt}{0pt}
\end{tabular}
}
\end{center}
\vspace{-4mm}
\caption{The mean and standard deviation of absolute size, relative size
and aspect ratio of objects in different datasets. Size is defined as the square root of the product of width and height and aspect ratio is the value of width divided by height. Absolute size is the size of object and relative size is the value of the object's size divided by the image's size. These settings follow as~\cite{Yu2020ScaleMF}.}
\label{tab: size of datasets}
\vspace{-13pt}
\end{table}

\subsubsection{Coarse Point Annotation}
\label{sec3:Coarse Point Annotation}
In practical scenarios, the coarse point can be obtained by annotating any single point on an object. However, since datasets in the experiment are already annotated with masks or bounding boxes, according to the law of large numbers, it is reasonable that the manually annotated points follow Gaussian distribution. Furthermore, since the annotated points must be inside the bounding box or mask of the object, 
then an improved Gaussian distribution, named Rectified Gaussian (RG) Distribution, is utilized for annotation. $RG(p;0, \frac{1}{4})$ is chosen to generate the point annotations for the experiments.

\vspace{-8pt}
\begin{equation}\small
\begin{aligned}
& \phi(p; \mu, \sigma) = Gauss(p;\mu, \sigma) \cdot Mask(p); \\
& RG(p; \mu, \sigma) = \frac{\phi(p;\mu, \sigma)} {\int_{p} \phi(p; \mu, \sigma)},
\label{Eq:point generation}
\end{aligned}
\vspace{-4pt}
\end{equation}
\vspace{-2pt}
where $\mu$ and $\sigma$ are the mean and standard deviation of the Gaussian distribution, respectively. $Mask(p)\in \{0, 1\}$ denotes whether point $p$ falls inside the mask of an object. If generated from the bounding box annotation, the box is treated as a mask.





\subsubsection{Implementation Details for CPRNet}
Our codes are based on MMDetection~\cite{mmdetection}. 
In dataset pre-processing, for COCO and VOC dataset, the short side of the images are resized to 400, and the ratio of width and height is kept. In DOTA dataset, images are split into sub-images (1024 $\times$ 1024 pixels) with overlap (200 $\times$ 200 pixels). And in SeaPerson, images are split into sub-images (640 $\times$ 640 pixels) with overlap (100 $\times$ 100 pixels). For data augmentation, only random horizontal flip is utilized in our CPRNet training.
ResNet-50 is used as the backbone network unless otherwise specified, and FPN is adopted for feature fusion.
P2 (stride is 4) is used for SeaPerson, and P3 (stride is 8) is used for COCO, VOC and DOTA. The mini-batch is 64/16/4/4 images, and all models are trained with 8/8/2/4 GPUs for COCO/VOC/DOTA/SeaPerson. Similar to the default setting of the object detection on COCO, the stochastic gradient descent (SGD ~\cite{DBLP:series/lncs/Bottou12})
algorithm is used to optimize. The training epoch numbers are set as 12/12/12/6, the learning rate is set as 0.00025, 0.0025, 0.00025, 0.0125 and decays by 0.1 at the 8-th/8-th/8-th/4-th and 11-th/11-th/11-th/5-th epoch for COCO/VOC/DOTA/SeaPerson, respectively. In default settings, the backbone is initialized with the pre-trained weights on ImageNet and other newly added layers are initialized with Xavier. In CPR, The sampling radius $R$ is set as 8/15/7/5 for COCO/VOC/DOTA/SeaPerson by default. And for CPR++, the init radius is set as 20/25/20/8 for COCO/VOC/DOTA/SeaPerson dataset.

\begin{table*}
\centering
    \begin{subtable}[t]{1.0\linewidth}
    \begin{center}
    \setlength{\tabcolsep}{9pt}
\resizebox{0.92\textwidth}{!}{
    \begin{tabular}{c|c|c|c|c|c|c|c|c}
    \toprule
        \multirow{2}{*}{Refinement} & \multirow{2}{*}{Localizer} & \multirow{2}{*}{Backbone} & \multicolumn{6}{c}{COCO~\cite{lin2014microsoft}} \\ 
        \cline{4-9}
         & & & ${\rm mAP}_{0.5}$ & ${\rm mAP}_{1.0}$ & ${\rm mAP}_{2.0}$ & ${\rm mAP}^{s}$  & ${\rm mAP}^{m}$ & ${\rm mAP}^{l}$ \\
        \midrule
        \midrule
        - & RetinaNet \cite{RetinaNet}   & ResNet-50 & 26.84 & 32.61 & 36.84 & 21.88 & 38.53 & 16.16 \\
        - & Faster-RCNN~\cite{DBLP:conf/nips/RenHGS15}  & ResNet-50 & 29.88 & 35.29 & 39.10 & 24.55 & 42.27 & 18.56 \\
        - & RepPoints~\cite{DBLP:conf/iccv/YangLHWL19}     & ResNet-50 & 31.76 & 37.42 & 41.51 & 25.26 & 43.92 & 20.22 \\
        - & P2PNet~\cite{Song_2021_ICCV}      & ResNet-50 & 32.14 & 38.85 & 43.54 & 23.94 & 37.81 & 19.09 \\
        \midrule
        Self-Refinement  & P2PNet~\cite{Song_2021_ICCV} & ResNet-50 & 44.11 &  51.06 & 55.61 & 18.80 & 51.38 & 41.56 \\
        CPR~\cite{DBLP:CPR}  & RetinaNet~\cite{RetinaNet}  & RestNet-50 & 44.82 & 51.35 & 55.68 & 19.39 & 50.68 & 47.55 \\
        CPR~\cite{DBLP:CPR}  & Faster-RCNN~\cite{DBLP:conf/nips/RenHGS15} & RestNet-50 & 47.29 & 53.21 & 56.85 & 20.20 & 52.56 & 50.61 \\
        CPR~\cite{DBLP:CPR}  & RepPoints~\cite{DBLP:conf/iccv/YangLHWL19} & RestNet-50 & 47.38 & 53.97 & 58.00 & 22.12 & 53.32 & 51.67 \\
        CPR~\cite{DBLP:CPR}  & P2PNet~\cite{Song_2021_ICCV} & ResNet-50 & 50.27 & 55.46 & 59.13 & 19.39 & 52.69 & 52.01 \\
        CPR~\cite{DBLP:CPR}  & P2PNet~\cite{Song_2021_ICCV} & ResNet-101 & 51.16 & 56.43 & 60.17 & 20.42 & 54.36 & 53.53 \\
        CPR++             & Faster-RCNN~\cite{DBLP:conf/nips/RenHGS15} & ResNet-50 & 49.99 & 55.36 & 58.99 & 19.43 &52.64 &57.17 \\
        CPR++             & RetinaNet~\cite{RetinaNet} & ResNet-50 & 47.63 & 53.80 & 58.02 & 18.82 & 50.44 & 54.74 \\
        CPR++             & RepPoints~\cite{DBLP:conf/iccv/YangLHWL19}  & ResNet-50 & 51.16 & 57.20 & 61.12 & 21.46 & 53.57 & 59.27 \\
        CPR++             & P2PNet~\cite{Song_2021_ICCV} & ResNet-50 & 53.06 & 58.22 & 61.80 & 19.77 & 53.48 & 59.13 \\
        
        CPR++             & P2PNet~\cite{Song_2021_ICCV} & ResNet-101 & 54.38 & 59.08 & 62.54 & 20.23 & 55.09 & 60.50 \\
        \bottomrule
        \end{tabular}}
        \end{center}
        \vspace{-8pt}
        \caption{The experimental comparison on COCO.}
        \label{tab:main coco}
    \vspace{5pt}
    \end{subtable}
    
    \begin{subtable}[t]{1.0\linewidth}
    \begin{center}
        \begin{tabular}{c|c|c|c|c|c|c|c|c|c|c}
        \specialrule{0.13em}{0pt}{0pt}
        \multirow{2}{*}{Refinement} & \multirow{2}{*}{Localizer} & \multicolumn{3}{|c|}{DOTA~\cite{DBLP:conf/cvpr/XiaBDZBLDPZ18}} & \multicolumn{3}{|c|}{SeaPerson} & \multicolumn{3}{|c}{VOC} \\ 
        \cline{3-11}
         & & ${\rm mAP}_{0.5}$  &${\rm mAP}_{1.0}$ & ${\rm mAP}_{2.0}$ & ${\rm mAP}_{0.5}$  &${\rm mAP}_{1.0}$ & ${\rm mAP}_{2.0}$ & ${\rm mAP}_{0.5}$  &${\rm mAP}_{1.0}$ & ${\rm mAP}_{2.0}$ \\
        \hline
        \hline
        - & RepPoints~\cite{DBLP:conf/iccv/YangLHWL19}       & 42.90 & 47.22 &50.11 & 40.53 & 47.15 & 54.20 & 41.75 & 49.84 & 53.39 \\
        - & P2PNet~\cite{Song_2021_ICCV}                    & 43.24 & 48.34 &51.60 & 72.52 & 76.80 & 80.15 & 44.03 & 52.57 & 56.12 \\
        \hline
        Self-Refinement & P2PNet~\cite{Song_2021_ICCV}      & 55.33 & 60.39 &63.22 & 79.97 & 85.19 & 87.57 &  43.16 & 53.06 & 58.41 \\ 
        CPR~\cite{DBLP:CPR} & P2PNet~\cite{Song_2021_ICCV}  & 60.10 & 63.81 &66.18 & 80.69 & 85.86 & 88.30 & 50.48 & 55.38 & 58.42 \\ 
        CPR++ & P2PNet~\cite{Song_2021_ICCV}                & 62.14 & 65.78 &67.92 & 81.55 & 86.24 & 88.64 & 51.69 & 56.19 & 58.71\\   
        \specialrule{0.13em}{0pt}{0pt}
        \end{tabular}
        \end{center}
        \vspace{-8pt}
        \caption{The experimental comparison on DOTA, SeaPerson and VOC.}
    \label{tab:other dataset}
    \end{subtable}
    \vspace{-5pt}
    \caption{The experimental comparisons (${\rm mAP}$) of localizers in four datasets: COCO, DOTA, SeaPerson and VOC. Localizer except P2PNet are trained with pseudo box (details in Sec.~\ref{sec:Detector with Pseudo Box}).}
    \label{tab:main}
    \vspace{-3pt}
\end{table*}


\begin{table*}[tb!]
\begin{center}

    \begin{subtable}[t]{0.3\linewidth}
    \begin{center}
    \begin{tabular}{l|c|c|c|c}
    \specialrule{0.13em}{0pt}{0pt}
    pos & MIL & ann & neg & ${\rm mAP}_{1.0}$ \\
    \hline
    \hline
    & \checkmark &            &            & 39.07  \\
    & \checkmark & \checkmark &            & 39.45 \\
    & \checkmark &            & \checkmark & 54.24 \\
    &           & \checkmark & \checkmark  & 51.82 \\
    \checkmark& & \checkmark & \checkmark  & 42.72 \\
    & \checkmark & \checkmark & \checkmark & \textbf{55.46} \\
    \specialrule{0.13em}{0pt}{0pt}
    \end{tabular}
    \end{center}
    \vspace{-5pt}
    \caption{Experiment of CPR loss.}
    \label{tab:training loss}
    \end{subtable}
    \begin{subtable}[t]{0.18\linewidth}
        \begin{tabular}{l|c|c}
        \specialrule{0.13em}{0pt}{0pt}
        feat & $R$ &${\rm mAP}_{1.0}$ \\
        \hline
        \hline
        & 5  & 53.32 \\
        & 8  & \textbf{55.46} \\ 
        P3 & 10 & 55.19 \\
        & 15 & 55.38 \\
        & 20 & 55.04 \\
        & 25 & 53.85 \\
        \specialrule{0.13em}{0pt}{0pt}
        \end{tabular}
        \caption{Different $R$ with P3.}
        \label{tab:sample_r P3}
    \end{subtable}
    \begin{subtable}[t]{0.18\linewidth}
        \begin{tabular}{l|c|c}
        \specialrule{0.13em}{0pt}{0pt}
        feat & $R$ &${\rm mAP}_{1.0}$ \\
        \hline
        \hline
        & 5 & 48.64 \\
        & 10  & 53.76 \\
        P2 & 15 & 54.26  \\
        & 20 & \textbf{54.64} \\
        & 30 & 54.24 \\
        & 40 & 53.11 \\
        \specialrule{0.13em}{0pt}{0pt}
        \end{tabular}
        \caption{Different $R$ with P2.}
        \label{tab:sample_r P2}
    \end{subtable}
    \begin{subtable}[t]{0.16\linewidth}
        \begin{center}
            \begin{tabular}{c|c}
            \specialrule{0.13em}{0pt}{0pt}
            shape & ${\rm mAP}_{1.0}$ \\
            \hline
            \hline
            $circle$      & 55.46 \\
            $rect._{1:1}$ & 55.39 \\
            $rect._{2:1}$ & 54.77 \\
            $rect._{1:2}$ & 54.96 \\
            \specialrule{0.13em}{0pt}{0pt}
            \end{tabular}
        \end{center}
        \caption{Sample shape.}
        \label{tab: sample selection shape}
    \end{subtable}
    \begin{subtable}[t]{0.15\linewidth}
        \begin{center}
            \begin{tabular}{c|c}
            \specialrule{0.13em}{0pt}{0pt}
            $u_{0}$ & ${\rm mAP}_{1.0}$               \\
            \hline
            \hline
            12 & 55.25 \\
            8  & 55.46 \\
            6  & 55.36 \\
            4  & 55.42 \\
            \specialrule{0.13em}{0pt}{0pt}
            \end{tabular}
        \end{center}
        \caption{Sample density.}
        \label{tab: sample denstity}
    \end{subtable}
\end{center}
    \vspace{-10pt}
\caption{Experiment of single CPR stage: (a) The effect of training loss in CPRNet: MIL loss, annotation loss, negative loss. The pos loss is for comparison. (b) and (c) are the performance with different $R$ on different feature map (P3, P2), where $R$ is the amount of sampling circles. (d) and (e) are the ablation studies on the sampling shape and sampling density.}
\vspace{-14pt}
\end{table*}

\subsection{Baseline Methods}
The POL task with coarse point annotation is divided into two key parts: refining the coarse point annotation and training the point localizer with refined points.


\vspace{-9pt}
\subsubsection{Detector with Pseudo Box} \label{sec:Detector with Pseudo Box} For training a point localizer, the intuitive idea is to convert the point-to-point (POL) to a box-to-box (object detection) problem. Firstly, a fixed-size pseudo-box is generated with each annotated point as the center. Next, the pseudo-box is used to train a detector. Finally, during inference, the center points of the boxes predicted by the trained detector are used as the final output. In this paper, following ~\cite{DBLP:conf/cvpr/RiberaGCD19}, we conduct the pseudo box for localization and give the performance in Table~\ref{tab:main}. On the COCO~\cite{lin2014microsoft} dataset, we give the performance of RetinaNet~\cite{RetinaNet}, FasterRCNN~\cite{DBLP:conf/nips/RenHGS15}, and RepPoint~\cite{DBLP:conf/iccv/YangLHWL19}. On the DOTA~\cite{DBLP:conf/cvpr/XiaBDZBLDPZ18} and SeaPerson datasets, we give the performance of RepPoint~\cite{DBLP:conf/iccv/YangLHWL19}.


%

\vspace{-9pt}
\subsubsection{Multi-Class P2PNet} \label{sec:p2p}
We adopt P2PNet\footnote{In our experiments, we re-implement P2PNet and further endow it the new ability of handling multi-class prediction, which to our best efforts, aligns the results with those reported in the raw paper~\cite{Song_2021_ICCV}.} as a
stronger baseline for POL tasks. It is trained with point annotation and predicts the point position for each object during inference. Improvement has been be made, especially when there are multiple categories:
\romannumeral1) The backbone of P2PNet in  this paper is Resnet-50 rather than VGG16~\cite{DBLP:journals/corr/vgg}. 
\romannumeral2) Instead of using the Cross-Entropy loss, we adopt the focal loss when optimizing classification to better deal with the problem of positive-negative imbalance; 
\romannumeral3) The Smooth-$\ell_1$ loss instead of $\ell_2$ loss is used for regression. 
\romannumeral4) In label assignment, different from a one-to-one matching in the default P2PNet, we assign top-k positive samples for each ground-truth and regard the remaining samples as background. And then, the NMS ~\cite{DBLP:conf/icpr/NeubeckG06} post-processing for points with fix-setting pseudo boxes is performed to obtain the final point results.
The performances of P2PNet improve a lot compared with the pseudo box on every dataset. P2PNet is a stronger baseline for POL tasks.

\vspace{-9pt}
\subsubsection{Self-Refinement} Inspired by ~\cite{DBLP:journals/corr/abs-1805-02641}, we propose a self-refinement technique that works as a strategy based on self-iterative learning for refining the coarse point annotation. Firstly, the aforementioned pseudo box strategy is adopted to train a point localizer. Then, the refined points are obtained through weighted mean of the points predicted by the localizer. 
Finally, the refined points are used as the new annotation to train the localizer. With these refined points as supervision, the performance of P2PNet as the point localizer is given in Table~\ref{tab:main}, indicating that it does alleviate the semantic variance problem to some extent.

 
\vspace{-10pt}
\subsection{Ablation Studies on CPR} \label{Sec: Ablation Studies on CPR}
To further analyze the effectiveness and robustness of single CPR stage, we conduct more experiments on COCO.


\vspace{-5pt}
\subsubsection{Training Loss in CPRNet}  
Ablation study of the training loss is given in Table~\ref{tab:training loss}. 
The CPR loss given in row 6 in~Table~\ref{tab:training loss} obtains 55.46 ${\rm mAP}$. \textbf{1) MIL loss.} If the MIL loss is removed (row 4), the CPRNet training relies on the annotation loss and the negative loss; the performance drops by 3.64 points (51.82 \emph{vs} 55.46). When we replace the MIL loss with the pos loss, which treats all the sampled points in the MIL bag as positive samples (line 5), the performance sharply declines by 12.74 points (42.72 \emph{vs} 55.46)
, showing that MIL can autonomously discern points belonging to the object. 
\textbf{2) Annotation loss.} Lacking of the annotation loss (row 3), the performance of localization decreases by 1.22 points (54.24 \emph{vs} 55.46). The annotation loss guides the training through a given accurate positive supervision.
\textbf{3) Negative loss.} 
With the negative loss (row 2), the performance improves by 16.01 points (55.46 \emph{vs} 39.45), indicating that only MIL loss is not enough to suppress the background, and the negative loss is inevitable.

\vspace{-6pt}
\subsubsection{Feature Map Level} 
The CPRNet is established based on a single level feature map of FPN. Table~\ref{tab:sample_r P3} and Table~\ref{tab:sample_r P2} show the performance with different feature map levels. Since the performance of P3 is similar to that of P2, P3 is chosen for our experiments in COCO if not otherwise specified.

\vspace{-5pt}
\subsubsection{Sampling Region}
 Table~\ref{tab:sample_r P3} and Table~\ref{tab:sample_r P2} show the performance of different radius $R$. It can be seen that $R$ is a sensitive hyper-parameter in CPRNet.  On P3, the best performance of 55.46 is obtained when $R$ is set as 8. If the sampling region reduces, such as $R=$ 5, the performance significantly declines to 53.32 since the sampling region is limited to a small local region, leading to a worse refinement.
While the region getting larger, the performance becomes steady but drops slowly until $R$ is over 25 (53.85) since the bag $B_j$ for MIL introduces more noise, which degrades the performance.

\subsubsection{Sampling Shape}
%
Theoretically, circles are orientation equivalent, which makes it suitable to model multi-view objects. Experimentally, the sampling results of the circle and rectangle are very close due to the discrete sampling, leading to comparable performance. In Table~\ref{tab: sample selection shape}, different aspect ratio $rect._{w:h}$ of rectangle sampling are also studied. 


\subsubsection{Sampling Density}
In Table~\ref{tab: sample denstity}, we ablate on sample density $u_0$. The performance is almost identical for different sample density from 4 to 12, which validates that $u_0$ has little impact.

\begin{table}[tb!]
\begin{center}
\begin{subtable}[t]{1.0\linewidth}
\centering
\setlength{\tabcolsep}{6.5pt}
\resizebox{0.9\textwidth}{!}{
    \begin{tabular}{c|c|c|c|c|c}
    \specialrule{0.13em}{0pt}{0pt}
    $R$ &${\rm mAP}_{1.0}$ & ${\rm mAP}^{s}$ & ${\rm mAP}^{m}$ & ${\rm mAP}^{l}$ & bag acc \\
    \hline
    \hline
    5  & 53.32 & 20.80 & 53.18 & 45.46 & 79.39 \\
    6  & 54.81 & 20.54 & 54.39 & 48.07 & 77.54 \\
    7  & 55.36 & 20.23 & 53.72 & 50.92 & 76.31 \\
    8  & 55.46 & 19.39 & 52.69 & 52.01 & 74.25 \\
    9  & 55.25 & 18.72 & 52.24 & 52.30 & 72.78 \\
    10 & 55.19 & 19.25 & 52.26 & 54.37 & 70.58 \\
    15 & 55.38 & 18.74 & 50.70 & 53.18 & 62.55 \\
    20 & 55.04 & 18.17 & 50.45 & 52.65 & 58.86 \\
    \specialrule{0.13em}{0pt}{0pt}
    \end{tabular}}
        \caption{Performance of different $R$ for CPR.}
    \label{tab: different $R$ for CPR}
\end{subtable}

\begin{subtable}[t]{1.0\linewidth}
\centering
\setlength{\tabcolsep}{7pt}
\resizebox{0.9\textwidth}{!}{
    \begin{tabular}{c|c|c|c|c|c}
    \specialrule{0.13em}{0pt}{0pt}
    iter & ${\rm mAP_{1.0}}$ & ${\rm mAP}^{s}$ & ${\rm mAP}^{m}$ & ${\rm mAP}^{l}$ & $Rate_{out}$\\
    \hline
    \hline
     0 & 38.85 & 23.94 & 37.81 & 19.09 & 0.04 \\
     1 & 55.46 & 19.97 & 53.57 & 51.81 & 0.14 \\
     2 & 55.84 & 17.24 & 52.99 & 56.59 & 0.20 \\
     3 & 55.17 & 16.84 & 52.01 & 57.08 & 0.22 \\
     4 & 55.27 & 17.76 & 51.91 & 57.67 & 0.24 \\
     5 & 55.11 & 16.42 & 52.35 & 57.85 & 0.25 \\
     6 & 55.13 & 16.22 & 51.88 & 59.21 & 0.26 \\
    \specialrule{0.13em}{0pt}{0pt}
    \end{tabular}}
    \caption{Performance of iterative CPR.}
    \label{tab:iterative CPR}
\end{subtable}
\end{center}
\vspace{-10pt}
\caption{Motivation of CPR++: minimal semantic variance.}
\vspace{-5pt}
\label{tab: motivation of cascade}
\end{table}

\vspace{-8pt}
\subsection{Experimental analysis on CPR++} \label{sec: Experimental analysis on CPR++}
\subsubsection{Motivation of CPR++}
In Table~\ref{tab: different $R$ for CPR}, we give the performance of large, medium and small objects for different sampling radius. Overall, the performance of small objects gradually decreases, and the performance of large objects gradually increases as the sampling radius increases. We believe that the main reason for this phenomenon is that the sampling region of CPR where finding the optimal solution is fixed. As the sampling radius increases, the noise (background points) also increases and the interference of adjacent objects is enhanced. The interference is also the reason for performance degradation when the sampling range is too large. 
In summary, a suitable sampling radius leads to performance improvement, but a continuous increase in sampling radius eventually leads to performance degradation. This inspired us to explore how to get a suitable sampling radius for the CPR network. 

In addition, we explored the cascade mode of the iterative CPR, which has been described in detail in Sec. \ref{sec: cascade designing}. As shown in Table~\ref{tab:iterative CPR}, iterating the CPR with a fixed small radius, the performance of the large objects continues to increase, and the performance of the small objects continues to decrease as the number of iterations increases, with a slight decrease in overall performance. 
We believe that the problem with this mode is due to the paradoxical nature of the optimization process. Since the sampling radius is fixed and the sampling center of each iteration is shifted, a positive sample in the previous iteration may become a negative sample in the next round, while a negative sample may become a positive sample. This will mislead the optimization in network training.

Combining Table~\ref{tab: different $R$ for CPR} and Table~\ref{tab:iterative CPR}, how to continuously optimize while ensuring high-quality label assignment becomes the key to the problem.

\begin{table}[tb!]
\begin{center}
\setlength{\tabcolsep}{2.5pt}
\resizebox{0.9\linewidth}{!}{
\begin{tabular}{c|c|c|c|c}
\specialrule{0.13em}{0pt}{0pt}
Method & ${\rm mAP}_{1.0}$ & ${\rm mAP}^{s}$ & ${\rm mAP}^{m}$ & ${\rm mAP}^{l}$ \\
\hline
\hline
CPR & 55.46 & 19.39 & 50.68 & 47.55 \\
Iterative CPR  & 55.84 & 17.24 & 52.99 & 56.59\\
Cascade CPR I \emph{w/o} var & 55.36 & 19.25 & 52.92 & 51.88 \\
Cascade CPR II \emph{w/o} var & 57.57 & 19.40 & 53.70 & 56.49 \\
Cascade CPR I  & 55.87 & 19.27 & 53.65 & 53.89 \\
Cascade CPR II (CPR++) & 58.22 & 19.77 & 53.48 & 59.13 \\
\specialrule{0.13em}{0pt}{0pt}
\end{tabular}}
\end{center}
\vspace{-10pt}
\caption{Experiment of various cascade mode. }\vspace{-5pt}
\label{tab:cascade mode}
\vspace{-6pt}
\end{table}


\begin{table}[tb!]
\begin{center}
\setlength{\tabcolsep}{9.5pt}
\resizebox{0.9\linewidth}{!}{
\begin{tabular}{c|c|c|c|c}
\specialrule{0.13em}{0pt}{0pt}
$K_s$  & ${\rm mAP}_{1.0}$ & ${\rm mAP}^{s}$ & ${\rm mAP}^{m}$ & ${\rm mAP}^{l}$\\
\hline
\hline
1 & 54.93 & 19.13 & 50.12 & 52.24 \\ 
2 & 57.57 & 19.40 & 53.70 & 56.49 \\
3 & 57.81 & 20.09 & 53.91 & 57.58 \\
\wj{4} & \wj{57.86} & \wj{19.46} & \wj{53.61} & \wj{58.62} \\ 
\specialrule{0.13em}{0pt}{0pt}
\end{tabular}}
\end{center}
\vspace{-10pt}
\caption{Ablation study of number of stages.}\vspace{-5pt}
\label{tab:number of stages}
\end{table}


\begin{table}[tb!]
\begin{center}
\setlength{\tabcolsep}{6.5pt}
\resizebox{0.9\linewidth}{!}{
\begin{tabular}{c|c|c|c|c|c}
\specialrule{0.13em}{0pt}{0pt}
$K_s$ & $L_{var}$ & ${\rm mAP}_{1.0}$ & ${\rm mAP}^{s}$ & ${\rm mAP}^{m}$ & ${\rm mAP}^{l}$\\
\hline
\hline
2      & & 57.57 & 19.40 & 53.70 & 56.49 \\
3      & & 57.81 & 20.09 & 53.91 & 57.58 \\
3      & \checkmark & 58.22 & 19.77 & 53.48 & 59.13 \\
\hline
2 & \checkmark & 55.68 & 19.42 & 51.95 & 53.42 \\
\specialrule{0.13em}{0pt}{0pt}
\end{tabular}}
\end{center}
\vspace{-10pt}
\caption{Ablation study of variance regularization.}
\label{tab:cascade_var_loss}
\vspace{-5pt}
\end{table}

\begin{table}[tb!]
\begin{center}
\setlength{\tabcolsep}{12pt}
\resizebox{0.9\linewidth}{!}{
\begin{tabular}{c|c|c|c}
\specialrule{0.13em}{0pt}{0pt}
Annotation & CPR & CPR++ & ${\rm mAP}_{1.0}$ \\
\hline
\hline
Coarse &  & & 38.48\\
Coarse & \checkmark & & 55.46 \\
Center & & & 57.47 \\
Coarse & & \checkmark & 58.22 \\
\specialrule{0.13em}{0pt}{0pt}
\end{tabular}}
\end{center}
\vspace{-10pt}
\caption{Comparisons with different annotation.}
\vspace{-15pt}
\label{tab:compare with center annotation.}
\end{table}


\vspace{-5pt}
\subsubsection{Different Cascade Modes}
Different cascade modes(Sec.~\ref{sec: cascade designing}) are ablated here to show the superiority of our newly designed cascade CPR II. In Table~\ref{tab:cascade mode}, we find that the performance gain of iterative CPR and cascade CPR I is limited. The performance of cascade CPR II achieves 58.22, which shows the effectiveness. Compared with single CPR, Cascade CPR II increases by 2.76 (58.22 vs 55.46), which greatly improves the existing state-of-the-art. Therefore, the cascade CPR II is adopted in CPR++.


\vspace{-5pt}
\subsubsection{Number of Stages}
Table~\ref{tab:number of stages} shows the effect of different numbers of stages on CPR++ performance. With only one stage, CPR++ obtains a performance of 54.93. As the number of stages increases, the performance of CPR++ gradually increases and reaches the best performance of 57.86 (average performance of experiments for three replicates) with four stages. Actually, the performance is convergent when three stages is utilized.
Therefore, three stages are chosen as our default configuration. 

\vspace{-5pt}
\subsubsection{Variance Loss}
In Table~\ref{tab:cascade_var_loss}, we further study the influence of variance loss: $L_{var}$. In our default three stage CPR++, the performance is 57.81. When variance loss is added, it increases to 58.22, which shows its effectiveness. We note that if adding var loss in two stage CPR++, the performance even decreases by about two points. This is because var loss aims to strengthen the feature of the foreground and reduce the spatial variance of the semantic points of the object representation. Only using two cascade stages, the noise of the solution space is still large. Multiple semantic features of the object are enhanced, the semantic variance increases and the network is misguided. Hence, the performance is even worse.



\vspace{-5pt}
\subsubsection{Comparison with Center Annotation} To further validate the CPR/CPR++, a comparison between CPR/CPR++ and a strict annotation based localizer, is conducted on COCO. 
Since it is hard to annotate the objects in the general dataset (\eg COCO) with key points, therefore, we approximately use the center point of each object's bounding box as a kind of strict point annotation. The experiment results in Tabel~\ref{tab:compare with center annotation.} show that CPR can achieve comparable performance to center point annotation based localizer (55.46 \emph{vs} 57.47). 
In CPR++, the localization performance exceeds that with center point annotation (58.22 \emph{vs} 57.47), showing that geometric center is not the best annotation. CPR++ can find a better semantic center point. 


\vspace{-10pt}
\subsection{Comparison with the State-of-the-Art Methods} \label{sec: Comparison with the State-of-the-Art}

\subsubsection{Performance comparison with baselines}
Considering that P2PNet has the highest performance compared with other localizers, we choose P2PNet as our baseline localizer. Meanwhile, the self-refinement strategy is chosen as the baseline method for coarse point refinement. Both of the two baselines have been stated in Sec.~\ref{sec:p2p}. In Table~\ref{tab:main coco}, compared with the baseline of P2PNet-ResNet-50, CPR and CPR++ achieve 16.61 and 19.37 performance improvement, respectively. Compared with the baseline of P2PNet with self-refinement strategy, CPR and CPR++ obtain 4.4 and 7.16 performance improvement, respectively. The performance of ${\rm mAP}_{0.5}$, ${\rm mAP}_{1.0}$ and ${\rm mAP}_{2.0}$ is reported for different strictness on COCO dataset. ${\rm mAP}^{s}$, ${\rm mAP}^{m}$ and ${\rm mAP}^{l}$ show the performance on objects of different scales. ${\rm mAP}_{1.0}$ is our core evaluation metric. The results of P2P train on coarse annotation and the points refined by CPR and CPR++ are visualized in Fig.~\ref{fig:visilization of P2P} and Fig.~\ref{fig:visilization of P2P 2} (different cases) in appendix.

\subsubsection{Different localizers}
We compare the performance of CPR and CPR++ on different localizers other than P2PNet. CPR and CPR++ achieve consistent and significant performance improvements compared to 3 localizers. CPR improves ${\rm mAP}_{1.0}$ by 18.74, 17.92, 16.55 and 11.42, CPR++ improves ${\rm mAP}_{1.0}$ by 21.19, 20.07 and 19.78 for RetinaNet, FasterRCNN and RepPoint respectively as shown in Table \ref{tab:main coco}. Furthermore, CPR++ obtains more performance improvements than CPR, with 2.45, 2.15 and 3.23 on three localizers.

\vspace{-5pt}
\subsubsection{Different backbones}
We further study the performance on different backbones in Table~\ref{tab:main coco}. CPR and CPR++ achieve 0.97 (56.43 vs. 55.46) and 0.86 (59.08 vs. 58.22) ${\rm mAP}_{1.0}$ performance gain when ResNet-50 is replaced by ResNet-101. CPR++ with ResNet-101 backbone and P2PNet localizer achieves the performance of 59.08 ${\rm mAP}_{1.0}$, which is State-of-the-Art.

\begin{table*}[]
    \centering
    \begin{tabular}{c|c|c|c|c|c|c}
    \hline
    \specialrule{0.13em}{0pt}{0pt}
     & \multicolumn{3}{c|}{ResNet 50} & \multicolumn{3}{|c}{ResNet 101} \\
     \hline
     & \multicolumn{2}{c|}{refiner} & localizer & \multicolumn{2}{c|}{refiner} & localizer \\
     & train & infer. & train/infer. & train & infer. & train/infer. \\
    \hline
CPR/CPR++(K=1) & 36.22 & 35.81 & 55.54 & 56.19 & 55.78 & 66.25 \\
CPR++(K=2) & 36.87 {\scriptsize(+1.8\%)} & 36.05 {\scriptsize(+0.7\%)} & 55.54 {\scriptsize(+0.0\%)} & 56.84 {\scriptsize(+1.2\%)} & 56.02 {\scriptsize(+0.4\%)} & 66.25 {\scriptsize(+0.0\%)} \\
CPR++(K=3) & 37.52 {\scriptsize(+3.6\%)} & 36.29 {\scriptsize(+1.3\%)} & 55.54 {\scriptsize(+0.0\%)} & 57.49 {\scriptsize(+2.3\%)} & 56.26 {\scriptsize(+0.9\%)} & 66.25 {\scriptsize(+0.0\%)} \\
CPR++(K=4) & 38.17 {\scriptsize(+5.4\%)} & 36.53 {\scriptsize(+2.0\%)} & 55.54 {\scriptsize(+0.0\%)} & 58.14 {\scriptsize(+3.5\%)} & 56.5 {\scriptsize(+1.3\%)} & 66.25 {\scriptsize(+0.0\%)} \\
CPR++(K=3) w. var & 38.3 {\scriptsize(+5.7\%)} & 36.29 {\scriptsize(+1.3\%)} & 55.54 {\scriptsize(+0.0\%)} & 58.27 {\scriptsize(+3.7\%)} & 56.26 {\scriptsize(+0.9\%)} & 66.25 {\scriptsize(+0.0\%)} \\
    \hline
    \specialrule{0.13em}{0pt}{0pt}
    \end{tabular}
    \vspace{1pt}
    \caption{\wj{The computation complexity (GFLOPs) for CPR and CPR++. The values in parentheses represent the percentage increase in computational load relative to CPR. (1) Each head in CPR++ adds only a classify branch and a instance branch, which are implemented with a linear layers. The computational load introduced by these components is not substantial. (2) During CPR/CPR++ inference, only the classification branch is utilized, results less computation increased. (3) CPR/CPR++ serves as a label refiner and does not affect the computational load of the localizer, which is the final component deployed in applications. (Since the increased computational load is related to the number of sampled points (radius), different samples may have different radius at different iterations during training due to CPR++'s dynamic radius. Therefore, here we estimate using the maximum possible radius for calculation, and the actual computation may be slightly lower than the values listed.)
    }}
    \label{tab: computation cost}
\end{table*}

\vspace{-5pt}
\subsubsection{Different datasets}
In addition to COCO, we explore the performance of CPR and CPR++ on DOTA and SeaPerson and VOC (Table~\ref{tab:other dataset}) to demonstrate the generalizability of CPR and CPR++.
On DOTA, CPR achieves 15.47 ${\rm mAP}_{1.0}$ performance improvement compared with the baseline localizer P2PNet, which overperforms the self-refinement strategy by 3.42 ${\rm mAP}_{1.0}$. CPR++ achieves a state-of-the-art performance of 65.78 ${\rm mAP}_{1.0}$ with 17.44 and 5.39 ${\rm mAP}_{1.0}$ performance gain compared with P2PNet and P2PNet with the self-refinement strategy.
On SeaPerson, CPR achieves 85.86 ${\rm mAP}_{1.0}$ performance, with 38.71, 9.06 and 0.67 point improvement compared with RepPoint, P2PNet and P2PNet with the self-refinement strategy. CPR++ overperforms CPR and reaches 86.24 ${\rm mAP}_{1.0}$, which is also State-of-the-Art. 
On VOC, CPR improves 5.54, 2.81 and 2.32 point compared with RepPoint, P2PNet and P2PNet with the self-refinement strategy. And CPR++ achieves 56.19, suppressed CPR.

\subsection{Analysis}
\subsubsection{Computation analysis}

\wj{We conducted a quantitative analysis of the computational overhead introduced by CPR++ in Table~\ref{tab: computation cost}. The analysis indicates that the additional computational load introduced by CPR++ is trivial. And there is no increased computational load for the localizer.}

\begin{table}[]
    \centering
    \begin{tabular}{c|c|c|c|c}
    \hline
    \specialrule{0.13em}{0pt}{0pt}
    & $\rm{mAP}$ & $\rm{mAP^s}$ & $\rm{mAP^m}$ & $\rm{mAP^l}$\\
    \hline
    CPR($R$=8)  & 55.46 & 19.39 & 52.69 & 52.01 \\
    CPR($R$=20) & 55.04 & 18.17 & 50.45 & 52.65 \\
    CPR++($R^{init}$=20)     & 58.22 & 19.77 & 53.48 & 59.13 \\
    \hline
    \specialrule{0.13em}{0pt}{0pt}
    \end{tabular}
    \vspace{1pt}
    \caption{\wj{Comparison between CPR++ and CPR with different sampling radius $R$.}}
    \label{tab: comparision CPR++ and CPR}
\end{table}

\subsubsection{Cross scale analysis}

\wj{In Table~\ref{tab:main coco}, the improvement in large object $\rm{mAP^l}$ is clearly more visible than the one in small and medium object ($\rm{mAP^s}$, $\rm{mAP^m}$). The main reason is that the larger the object, the more pronounced the semantic variance, and the more evident the effects of CPR/CPR++. Both CPR and CPR++ methodologies are tailored to mitigate the challenges associated with semantic variance. As delineated in Fig. 1(a)(b), semantic variance essentially stems from the ambiguity in training due to the semantic distinctions among annotated points. Larger objects tend to exhibit a higher count of annotatable points and a more intricate semantic structure, exacerbating the semantic variance issue. In essence, larger objects provide a more substantial scope for enhancement, leading to a more profound impact for CPR/CPR++ on larger objects. In addition, the performance of CPR in Table~\ref{tab:main coco} is actually CPR with small sampling radius $R$=8, which is advantageous for small objects. When the CPR radius is increased, it obtains better performance on large objects, but the performance on small objects tends to decline as shown in Table~\ref{tab: comparision CPR++ and CPR}. The initial radius $R^{init}$ of CPR++ is also 20. In comparison to CPR (R=20), CPR++ exhibits more noticeable improvements on small and medium-sized objects than CPR (R=8) as shown in Table~\ref{tab:main coco}.}



\vspace{-5pt}
\section{Conclusion}
In this work, we rethink the semantic variance problem in point-based annotation caused by the non-uniqueness of optional annotated points. The proposed CPR samples points in the neighborhood and then detects the semantic points of each object by using MIL. 
The weighted average of the semantic points is the semantic center of the object, which is used as the supervision for learning the localizer. CPR alleviates semantic variance and facilitates the extension of POL tasks to multi-class and multi-scale. Furthermore, CPR++ is introduced to further handle the multi-scale problem and reduce the semantic variance. 
Extensive experiments on various datasets validate the effectiveness of our methods.


%


\vspace{-10pt}
\ifCLASSOPTIONcompsoc
  \section*{Acknowledgments}
\else
  \section*{Acknowledgment}
\fi

This work was supported in part by the Youth Innovation Promotion Association
CAS, the National Natural Science Foundation of China (NSFC) under Grant No.
61836012 and 61771447, the Strategic Priority Research Program of the
Chinese Academy of Sciences under Grant No.XDA27000000.

\ifCLASSOPTIONcaptionsoff
  \newpage
\fi



\small
\bibliographystyle{IEEEtran}
\bibliography{IEEEabrv,egbib}

\vspace{-13mm}

\begin{IEEEbiography}[{\includegraphics[width=1in,height=1.25in,clip]{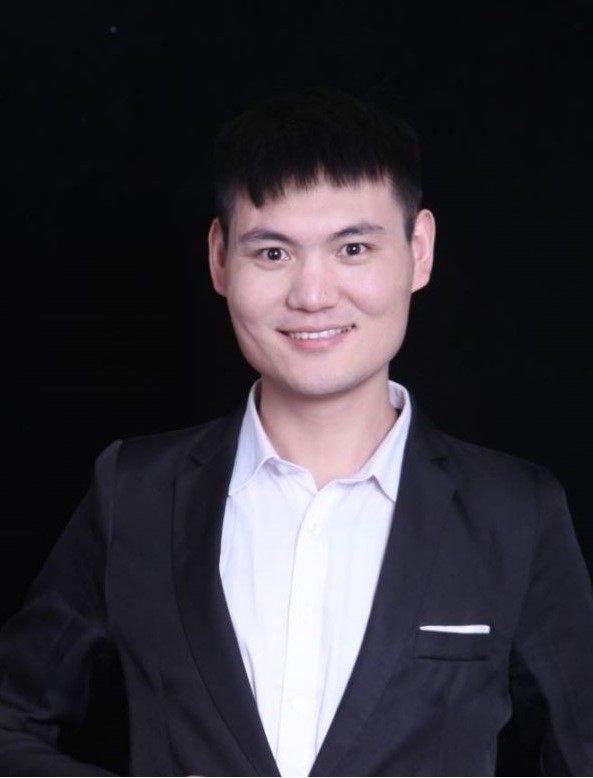}}]{Xuehui Yu}
received the B.E. degree in software engineering from Tianjin University, China, in 2017. He is currently pursuing the Ph.D. degree in signal and information processing with the School of Electronic, Electrical, and Communication Engine, University of Chinese Academy of Sciences. His research interests include machine learning and computer vision.
\end{IEEEbiography}
\vspace{-13mm}
\begin{IEEEbiography}[{\includegraphics[width=1in,height=1.25in,clip]{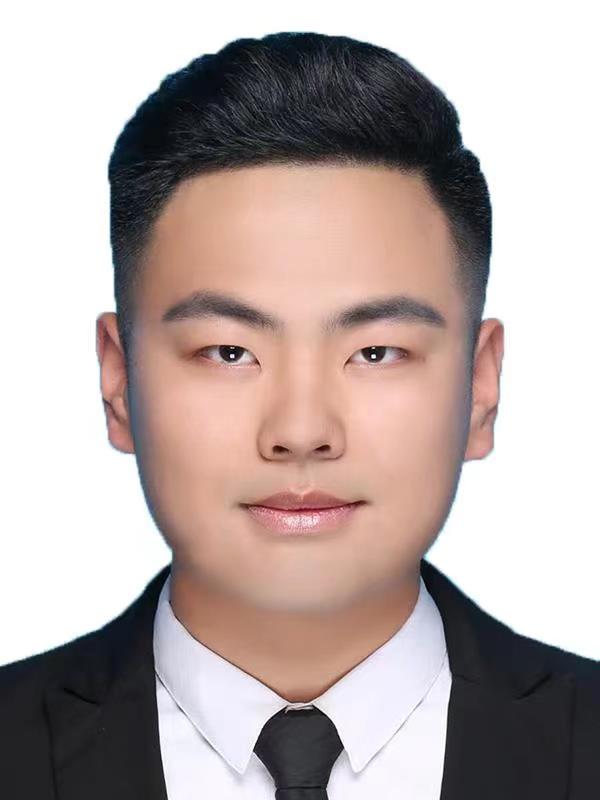}}]{Pengfei Chen}
received the B.E. degree in electronic and information engineering from Civil Aviation University of China, China, in 2020. He is currently pursuing the Ph.D. degree in signal and information processing with the School of Electronic, Electrical, and Communication Engine, University of Chinese Academy of Sciences. His research interests include machine learning and computer vision.
\end{IEEEbiography}

\begin{IEEEbiography}[{\includegraphics[width=1in,height=1.25in,clip]{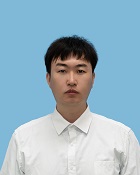}}]{Kuiran Wang}
 received the B.E. degree in computer science and technology from Central South University, China, in 2018. He is currently pursuing the Ph.D. degree in electronic and communication engineering with the School of Electronic, Electrical, and Communication Engine, University of Chinese Academy of Sciences. His research interests include machine learning and computer vision.
\end{IEEEbiography}
\begin{IEEEbiography}[{\includegraphics[width=1in,height=1.25in,clip]{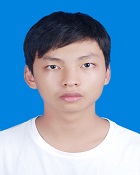}}]{Xumeng Han}
received the B.E. degree in electronic and information engineering from University of Electronic Science and Technology of China, Chengdu, China, in 2019. He is currently pursuing the Ph.D. degree in signal and information processing with the School of Electronic, Electrical, and Communication Engine, University of Chinese Academy of Sciences, Beijing, China. His research interests include machine learning and computer vision.
\end{IEEEbiography}


\begin{IEEEbiography}[{\includegraphics[width=1in,height=1.25in,clip]{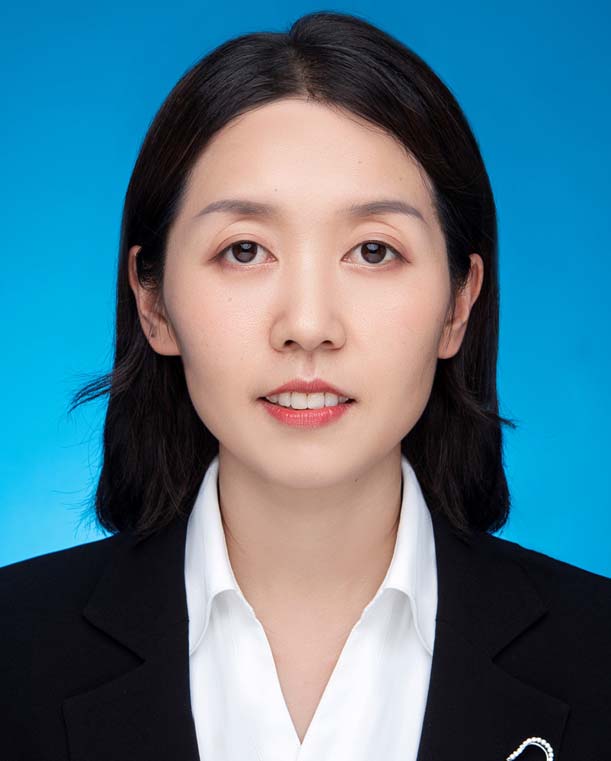}}]{Guorong Li}
received her B.S. degree in technology of computer application from Renmin University of China, in 2006 and Ph.D. degree in technology of computer application from the Graduate University of the Chinese Academy of Sciences in 2012. Now, she is an associate professor at the University of Chinese Academy of Sciences. Her research interests include object tracking, video analysis, pattern recognition, and cross-media analysis.
\end{IEEEbiography}


\begin{IEEEbiography}[{\includegraphics[width=1in,height=1.25in,clip]{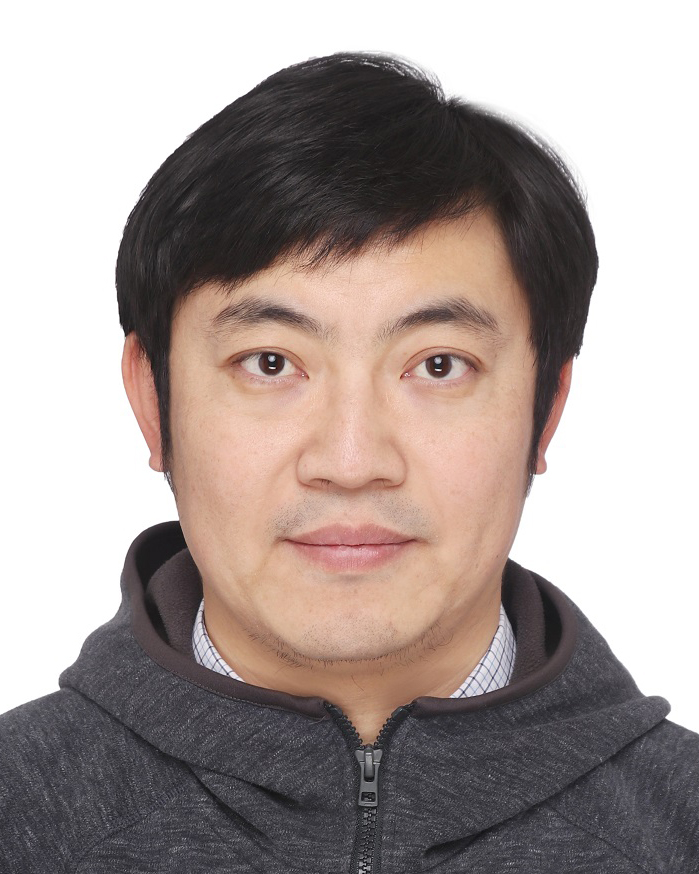}}]{Zhenjun Han}
received the B.S. degree in software engineering from Tianjin University, Tianjin, China, in 2006 and the M.S. and Ph.D. degrees from University of Chinese Academy of Sciences, Beijing, China, in 2009 and 2012, respectively. Since 2013, he has been an Associate Professor with the School of Electronic, Electrical, and Communication Engineering, University of Chinese Academy of Sciences. His research interests include object tracking and detection.
\end{IEEEbiography}


\begin{IEEEbiography}[{\includegraphics[width=1in,height=1.25in,clip]{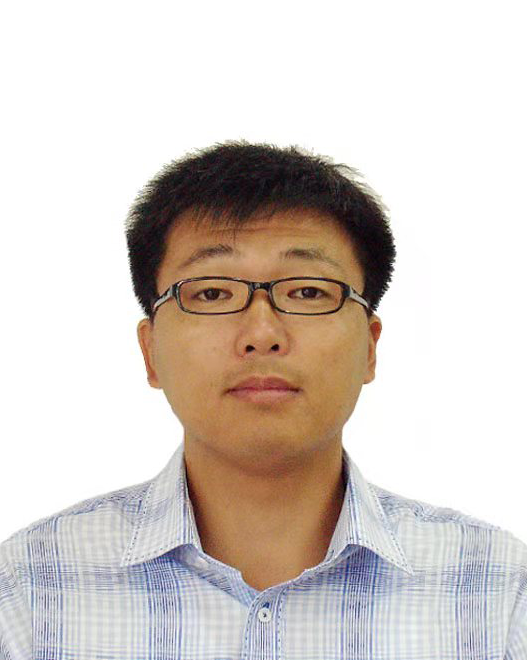}}]{Qixiang Ye}
received the B.S. and M.S. degrees from Harbin Institute of Technology, China, in 1999 and 2001, respectively, and the Ph.D. degree from the Institute of Computing Technology, Chinese Academy of Sciences in 2006. He has been a professor with the University of Chinese Academy of Sciences since 2009, and was a visiting assistant professor with the Institute of Advanced Computer Studies (UMIACS), University of Maryland, College Park until 2013. His research interests include image processing, visual object detection and machine learning. He has published more than 100 papers in refereed conferences and journals including IEEE CVPR, ICCV, ECCV and PAMI. He is on the editorial boards of IEEE Transactions on Intelligent Transportation System and IEEE Transactions on Circuit and System on Video Technology.
\end{IEEEbiography}

\vspace{-400pt}
\begin{IEEEbiography}[{\includegraphics[width=1in,height=1.25in,clip]{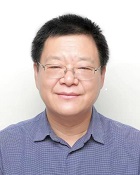}}]{Jianbin Jiao}
(Member, IEEE) received the B.S., M.S., and Ph.D. degrees in mechanical and electronic engineering from the Harbin Institute of Technology (HIT), Harbin, China, in 1989, 1992, and 1995, respectively. From 1997 to 2005, he was an Associate Professor with HIT. Since 2006, he has been a Professor with the University of Chinese Academy of Sciences, Beijing, China. His research interests include computer vision and pattern recognition.
\end{IEEEbiography}

\newpage

\appendix

\textbf{1. Visualization of P2P}
\vspace{5pt}

\begin{figure*}[t!]
  \begin{subfigure}{1.0\linewidth}
    \centering
    \includegraphics[width=0.95\linewidth]{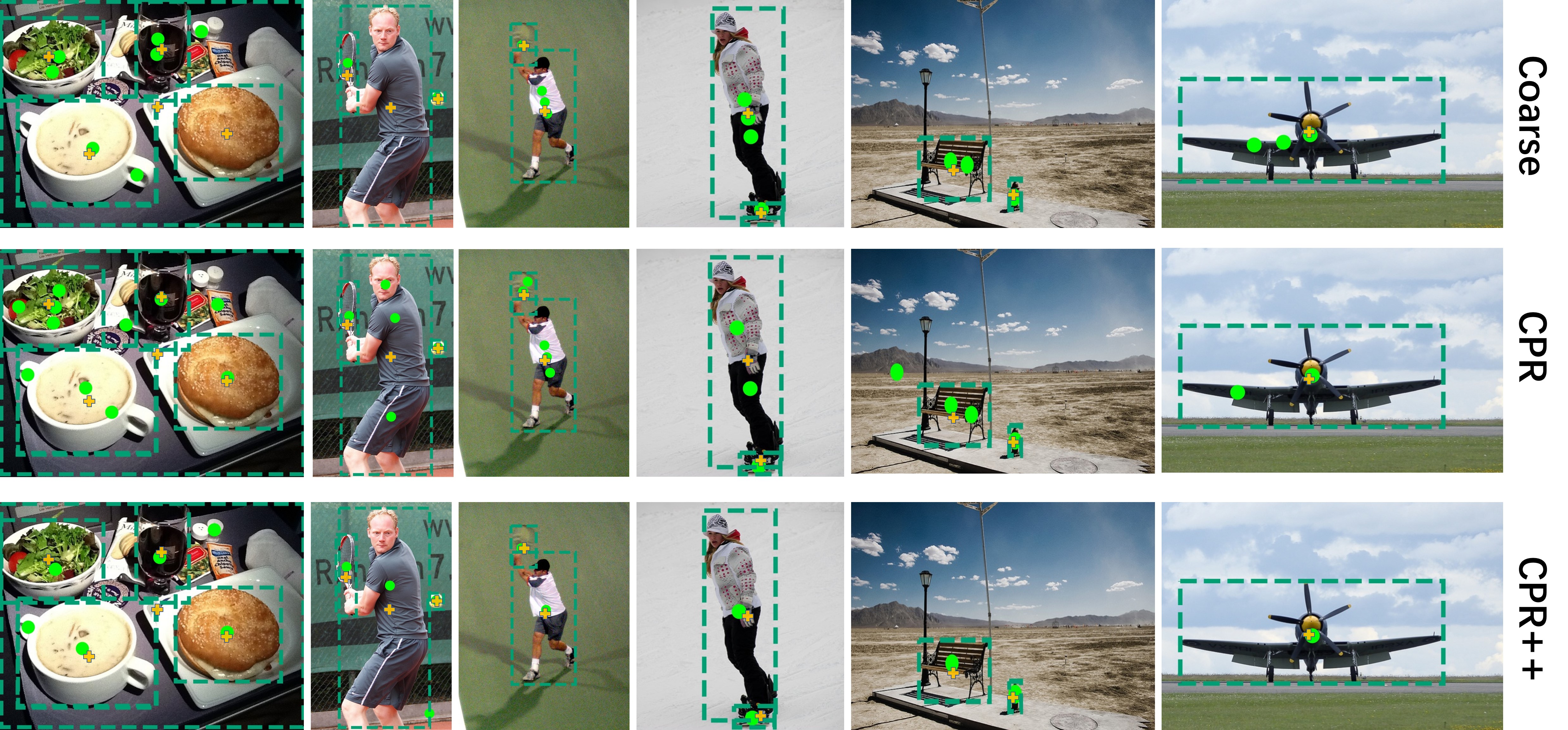}
  \end{subfigure}
  \vspace{-5pt}
  \caption{Visualization of p2p with different supervision: original coarse annotation, refined point of CPR, refined point of CPR++. The first row shows the results of a localizer trained directly using coarse points as supervision. The second/third row illustrates the results of a localizer trained with the points refined by CPR/CPR++ from the coarse points, respectively. It can be observed that networks trained with either coarse points or CPR refined points tend to have more instances lost or more false positives comparing with the network trained with CPR++ refined points. "lost" indicates that some objects do not have any predicted localization points on them, while "false positive" indicates that some objects have multiple localization points on them.  
  \wjj{The green dash bounding box and orange cross symbol is the box-level ground-truth and center point of ground-truth box for better view.}
  }
  \vspace{-15pt}
\label{fig:visilization of P2P}
\end{figure*}

\wj{Fig.~\ref{fig:visilization of P2P} and Fig.~\ref{fig:visilization of P2P 2} display the inference results of P2P localizer trained with different types of points as supervision. The boxes in these Figs are the box-level ground-truth. The points are the network predictions. Each point indicates a predicted object.
In Fig.~\ref{fig:visilization of P2P 2}, for object in different cases, compared to the original coarse point annotations, CPR has lower false positives due to the decrease of the semantic variance. Since the semantic variance of CPR++ further decreases, due to the more global solution space, and then less false positives are obtained (less green points).}

\vspace{5pt}
\noindent \textbf{2. Failure analysis of the proposed CPR++}
\vspace{5pt}

\wjj{
To identify potential improvements for CPR++, we visualized and analyzed failure cases in Fig~\ref{fig: visilization of failed case}. These failure cases can be broadly categorized into four types: 1) dense and occlusion 2) huallucination from glass reflections.
3) blur and small 4) incorrect annotations in the COCO dataset.}

\vspace{5pt}
\noindent \textbf{3. Trade-off between performance and computation cost}
\vspace{5pt}

\begin{figure}[]
    \centering
    \includegraphics[width=1.0\linewidth]{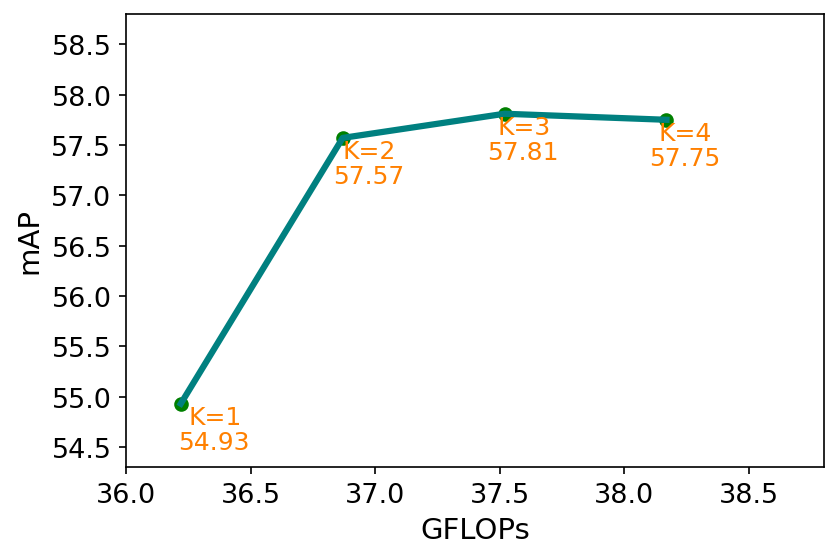}
  \vspace{-5pt}
  \caption{The performance and computation cost for CPR with different number of heads.}
  \vspace{-15pt}
\label{fig:trade-off}
\end{figure}

\begin{figure*}[tb!]
  \hfill
  \begin{subfigure}{1.0\linewidth}
    \centering
    \includegraphics[width=0.95\linewidth]{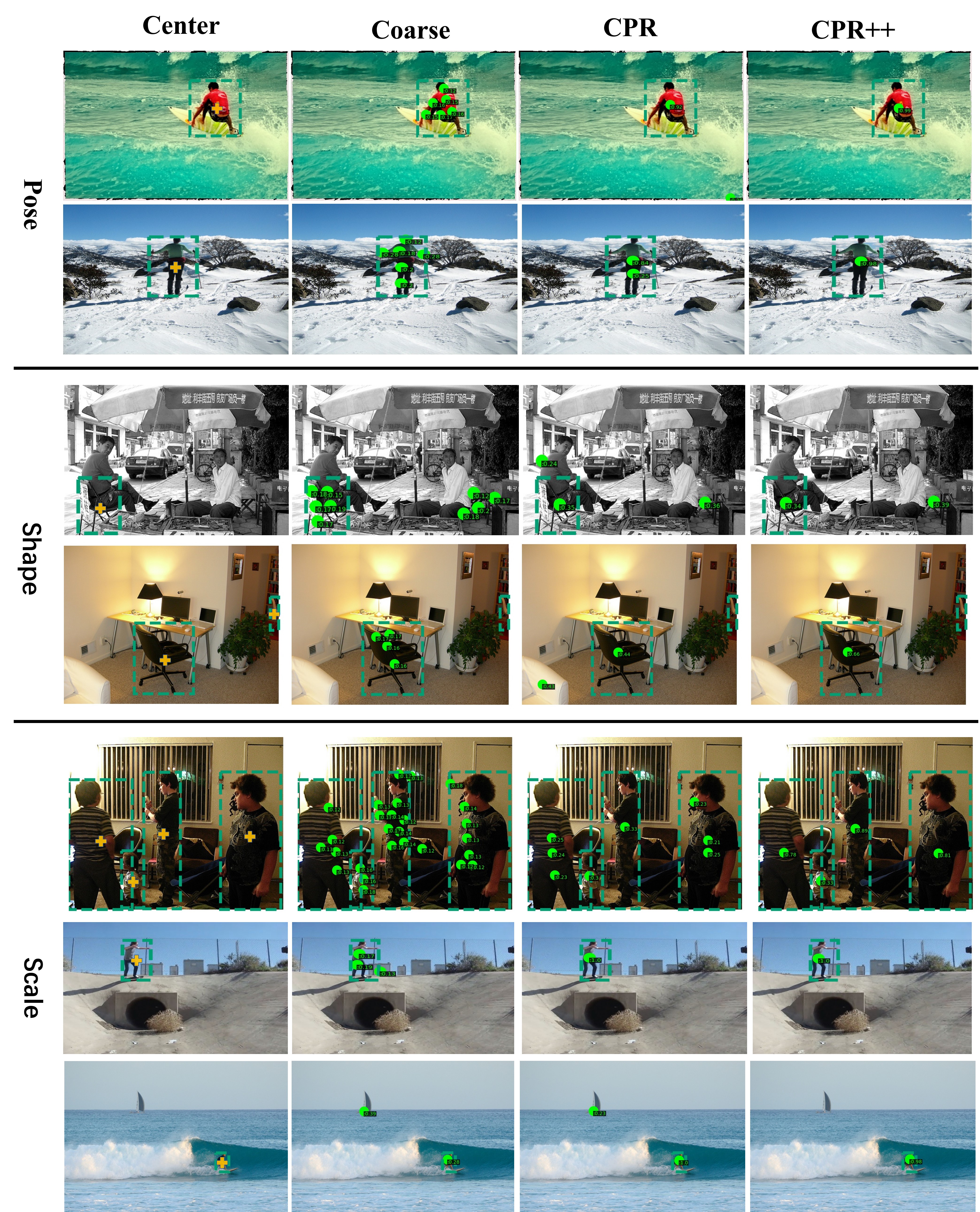}
  \end{subfigure}
  \vspace{-5pt}
  \caption{Visualization of method comparison in different cases. The dash boxes are box-level ground-truth, neither annotations nor network predicted results. The points are the network predicted results. Each point indicates a predicted object. \wjj{The column "center" represents the center point (orange cross symbol) of ground-truth box.} The column “coarse” represents the predicted results of the state-of-the-art localizer P2PNet that directly trained with the original annotations. "CPR" and "CPR++" indicate the visualization results of the P2PNet trained with the points refined by CPR and CPR++, respectively. For object in different cases, compared to the original coarse point annotations, CPR has lower false positives due to the decrease of the semantic variance. Since the semantic variance of CPR++ further decreases, due to the more global solution space, and then less false positives are obtained (less green points).
}
  \vspace{-15pt}
\label{fig:visilization of P2P 2}
\end{figure*}

\begin{figure*}[tb!]
  \hfill
  \begin{subfigure}{1.0\linewidth}
    \centering
    \includegraphics[width=0.95\linewidth]{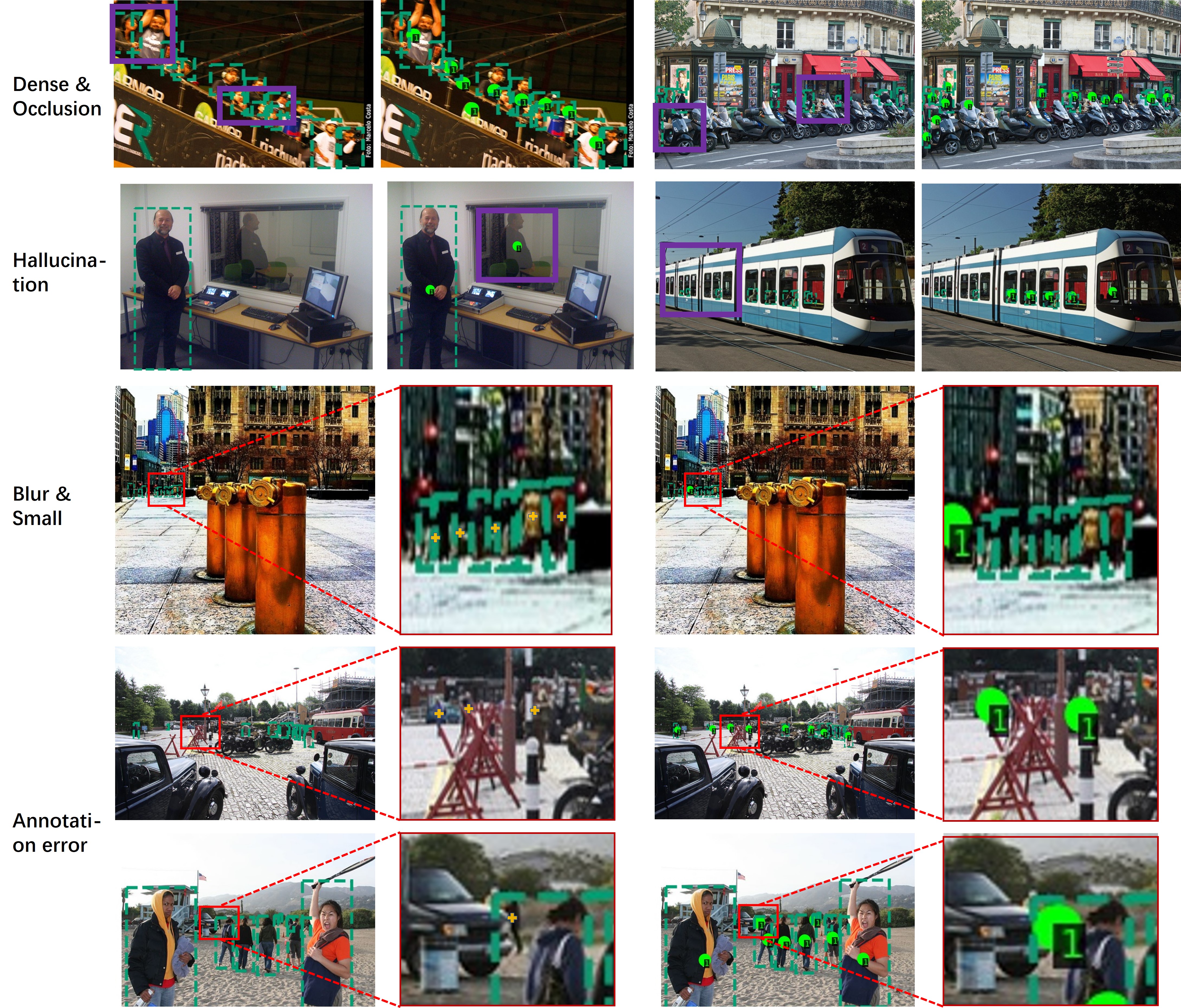}
  \end{subfigure}
  \vspace{-5pt}
  \caption{\wjj{Visualization of failed cases of CPR++. The green point and dash green box are detection results of CPR++ and box-level ground truth, respectively. For the last 3 rows, we enlarge the region in the red box for better presentation. 
  \textbf{Dense \& Occlusion}: as with all visual tasks, scenes with extremely dense objects pose a significant challenge. This is primarily due to the substantial occlusion caused by the density of the objects. In such scenarios, CPR++ may exhibit false positives and false negatives, as illustrated by the purple boxes in the first row.
  \textbf{Hallucination}: in the COCO dataset, the person seated inside vehicles (behind glass) and some characters reflected in mirrors or TVs are annotated as "person." This can lead to hallucinations in the network's recognition of inside vehicles and reflections.
  \textbf{Blur \& Small}: CPR++ tends to miss some very small and blurry objects. (objects labelled with an orange cross in the 3rd row)
  \textbf{Annotation error}: CPR++ demonstrates the capability to identify some instances that may have been missed during COCO annotation.  (objects labelled with an orange cross in the 4-th and 5-th rows)}
}
  \vspace{-15pt}
\label{fig: visilization of failed case}
\end{figure*}

\wj{To reduce this increased computation cost, we can stop iterating the refined stage for some poorly refined objects or objects that do not show significant changes (convergence) between two adjacent stages, rather than applying the K stage to all objects. This strategy can reduce certain computations to some extent. In addition, Fig.~\ref{fig:trade-off} shows the computational load and performance with K heads in CPR++. When K is 1, it is equivalent to the original CPR. When K is set as 2, adding one head to CPR, obtains a significant performance improvement (+2.64), since the addition of the first head represents a substantial change from CPR with a fixed sampling radius to a dynamic sampling radius. While the performance improvement is limited (+0.29) when K is larger than 2 since subsequent head additions mainly provide fine-grained adjustments to the dynamic radius. Therefore, we can consider CPR(K=2) as a good trade-off between performance and computational complexity.}

\end{document}